\documentclass{article}

\usepackage[authoryear,longnamesfirst]{natbib}
\usepackage{caption}
\usepackage{subcaption}
\usepackage[english]{babel}
\usepackage{algorithm}
\usepackage{algpseudocode}
\usepackage{amsmath}
\usepackage{geometry}

\usepackage{graphicx}
\usepackage{hyperref}

\usepackage[utf8]{inputenc}
\usepackage{pgfplots}
\DeclareUnicodeCharacter{2212}{−}
\usepgfplotslibrary{groupplots,dateplot}
\usetikzlibrary{patterns,shapes.arrows}
\pgfplotsset{compat=newest}
\usepackage{longtable}
\usepackage{multirow}
\usepackage{afterpage}  
\usepackage[noblocks]{authblk}

\title{Elastic Step DQN: A novel multi-step algorithm to alleviate overestimation in Deep Q-Networks}             

\author[1]{Adrian Ly}
\affil[1, 2, 5]{\texttt{Deakin University}}
\author[2]{Richard Dazeley}
\author[3]{Peter Vamplew}
\affil[3]{\texttt{Federation University}}
\author[4]{Francisco Cruz}
\affil[4]{\texttt{UNSW}}
\author[5]{Sunil Aryal}
\date{}

\begin{document}

\maketitle

\begin{abstract}

Deep Q-Networks algorithm (DQN) was the first reinforcement learning algorithm using deep neural network to successfully surpass human level performance in a number of Atari learning environments. However, divergent and unstable behaviour have been long standing issues in DQNs. The unstable behaviour is often characterised by overestimation  in the $Q$-values, commonly referred to as the overestimation  bias. To address the overestimation  bias and the divergent behaviour, a number of heuristic extensions have been proposed. Notably, multi-step updates have been shown to drastically reduce unstable behaviour while improving agent's training performance. However, agents are often highly sensitive to the selection of the multi-step update horizon ($n$), and our empirical experiments show that a poorly chosen static value for $n$ can in many cases lead to worse performance than single-step DQN. Inspired by the success of $n$-step DQN and the effects that multi-step updates have on overestimation  bias, this paper proposes a new algorithm that we call `Elastic Step DQN' (ES-DQN). It dynamically varies the step size horizon in multi-step updates based on the similarity of states visited. Our empirical evaluation shows that ES-DQN out-performs $n$-step with fixed $n$ updates, Double DQN and Average DQN in several OpenAI Gym environments while at the same time alleviating the overestimation  bias. 

\end{abstract}

\newpage
\section{Introduction}

 One of the most prominent reinforcement learning algorithms is Q-learning \citep{watkins1992q}, which has been shown to converge to optimal policies when used with lookup tables. However, tabular Q-learning is limited to small sized and toy environments, and lookup tables are computationally inefficient when environments have large state action spaces. To address the scalability of Q-learning, deep neural networks became a viable alternative to lookup tables to approximate state-action values in large continuous spaces. One of the most popular algorithms is the Deep Q-Network (DQN) \citep{mnih2013playing, mnih2015human} which uses neural networks and introduced the concept of a target network and replay memory with great success in the Atari games environment. In spite of the success, training instability and divergent behaviour is regularly observed in DQN \citep{sutton2018reinforcement, van2016deep}. The divergent behaviour itself was not specific to DQN, and has also been heavily investigated for function approximators in the past \citep{sutton2018reinforcement, baird1995residual, tsitsiklis1997analysis} with a number of linear solutions having been proposed to address the problem \citep{maei2009convergent, baird1995residual}. Overestimation  of the $Q$-values has frequently been identified as one of the key reasons that causes sub-optimal learning and divergent behaviour -- this was thoroughly investigated by \citet{thrun1993issues} who attributed the issue to noise generated through function approximation. On the other hand, \citet{hasselt2010double} theorised that the overestimation  originated from the max operator used as part $Q$ value updates which tends estimations towards larger values and a more optimistic outlook. \citet{van2018deep} and \citet{van2016deep} have suggested that by correcting for overestimation, an agent is much less susceptible to divergent behaviours. 

Empirical observations by \citet{van2018deep} and \citet{hessel2018rainbow} characterised a number of plausible architectural and algorithmic mechanics that may lead to divergent behaviour as well as suggestions that may alleviate divergence. 
\citet{van2016deep} also hypothesised that multi-step returns are likely to reduce the occurrence of divergence. This idea that multi-step DQN updates can regulate and reduce divergent behaviour while at the same time return stronger training performance is not without merit. Multi-step implementations like the Rainbow agent \citep{hessel2018rainbow} and Mixed Multi-step DDPG \citep{meng2021effect} have shown empirically that multi-step updates under certain conditions are more stable, and in many circumstances can circumvent the divergence problem that exists in single-step DQN. However, studies have also shown that multi-step DQN updates are highly sensitive to the selection of the value $n$ which if incorrectly selected can be detrimental to learning \citep{hessel2018rainbow, chiang2020mixture, horgan2018distributed, deng2020value, fedus2020revisiting}. This raises the question then, is a static value of $n$ the best approach to multi-step updates or is it possible to dynamically select the $n$ parameter and take advantage of the special properties that multi-step updates provide.  

One possible approach is inspired by the work of \citet{dazeley2015coarse}, who identified an issue where agents may diverge in a grid world setting under linear approximation due to a self-referential learning loop problem. It was suggested that the consolidation of consecutive states and the treatment of them as sub-states of larger state, can encourage more stable algorithmic performance. However the ideas were limited to a grid world setting under linear function approximation. In this paper, we extend the ideas of \citet{dazeley2015coarse} into a deep reinforcement learning context by leveraging a multi-step update. 

As a combination of these ideas, this paper will introduce Elastic Step DQN (ES-DQN) -- a novel multi-step update that dynamically selects the step size horizon based on consecutive state similarity. The intuition behind the idea of consolidating similar states together is in general inspired by spatial information theory which identified that people have a natural inclination to use landmarks to provide route directions \citep{michon2001and, klippel2005structural}. When providing directions, individuals tend to summarise the instructions based on broader objectives and use landmarks to anchor and help generate mental images (e.g. `walk down the end of the hallway and turn left once you see the toilet', in comparison to more detailed step by step descriptions -- `take one step then another step until you reach the end of the hallway'). Our experimental results indicate that the systematic accumulation of experiences that are similar is able to achieve comparable performance to the well established DQN extensions and superior performance to $n$-step updates. Furthermore to achieve statistical significance for the experiments, this paper will primarily use three small sized environments as its test bed for this investigation. 

The following sections will investigate the efficacy of multi-step updates, against single-step DQN as well as introduce a novel multi-step DQN update that automates the selection of the step size horizon. The algorithm will also be compared against DoubleDQN and Average DQN which are well established methods of reducing overestimation. This paper makes four key contributions: \begin{itemize}
    \item introduce the idea of dynamic multi-step updates based on state similarity
    \item the introduction of ES-DQN to select the $n$ step horizon dynamically
    \item an in-depth empirical exploration of $n$-step DQN against ES-DQN
    \item and an in-depth empirical analysis of algorithmic stability when grouping together similar states. 
\end{itemize}

\section{Related works}

\subsection{Background} \label{sec:background}

The training of a reinforcement learning agent is typically unguided and is characterised by a Markov decision process where the agent would at each time step $t$ observe a particular state $S_t$ before it takes an action $A_t$. The agent would receive some sort of reward $R_{t+1}$ and enter into a new state $S_{t+1}$ -- the main objective is for the agent to learn an optimal policy that maximises the expected discounted accumulation of rewards $G_{t}$ at time $t$ to $k$ steps into the future, at the discount factor $\gamma$. 
\begin{displaymath}
\centering
G_t = \sum_{k=1}^{k-1} \gamma^{k}_{t} R_{t+k+1}   
\end{displaymath}

Classical reinforcement learning algorithms rely largely on lookup tables to store state action values to learn an optimal policy -- one such algorithm is Q-learning which was first introduced by \citep{watkins1992q} and has since been a focal point of value based reinforcement learning approaches. As a Q-learning agent navigates its environment and receives responses from the environment, the lookup table is updated at every time step where $Q$ is the state-action value function, $\alpha$ is the learning rate, $R$ is the reward, and $A^{'}$ is an action selected from a policy derived from $Q$ \cite{sutton2018reinforcement}. 

\begin{displaymath}
\centering
Q^{new}(S, A)\gets Q(S, A)+\alpha[ \,R + \gamma \max_{A^{'}}Q(S^{'},  A^{'}) - Q(S, A)]
\end{displaymath}

Multi-step updates \citep{sutton2018reinforcement} expand on single-step by temporally extending the step size horizon of the target by the value of $n$, traditionally multi-step updates were infrequently used because it was inconvenient to implement and were more used as an intermediate step to eligibility traces. However, multi-step updates have shown to allow for faster convergence \citep{sutton2018reinforcement, ceron2021revisiting} when the value of $n$ is well-tuned.
\subsection{Related literature} \label{sec:literature}

Conceptually the proposed multi-step updates in this paper extends on the ideas introduced in Options \citep{sutton1999between} and Coarse Q-learning \citep{dazeley2015coarse}. Options were first formalized as a framework to allow agents to take sub-actions under a single policy until a termination condition is activated. The termination condition can come in the form of a sub-goal being achieved or when a certain temporal range is met. Similarly Coarse Q-learning builds on the Options framework and explores the issue of a wandering agent in a coarsely quantised environment. \citet{dazeley2015coarse} showed that divergence under quantisation can be explained by a wandering phenomenon where an agent when trapped in the same state for multiple updates may wander aimlessly and exhibit unstable learning. To address the behaviour, \citet{dazeley2015coarse} introduces clamped actions which force the agent to take the same action until the quantised state is exited, and the $Q$-function is only updated once this occurs. While conceptually both Options and Coarse Q-learning is similar to our approach, both frameworks extend largely to linear function approximators.  

Interestingly, while our idea is relatively simple, the concept of consolidating experiences based on state similarity to address value-based divergence has been relatively unexplored in deep reinforcement learning. 
Early works like the Rainbow agent incorporated multi-step updates into DQN \citep{hessel2018rainbow, van2018deep} which surpassed super-human performance benchmarks set by the DQN and Double DQN agents. 
While the study indicated that the incorporation of multi-step methods had a strong positive impact on overall agent performance, it was limited in sample size due to the computational intensity of the Atari games. \citet{ceron2021revisiting} further investigated this with a large exploratory set of experiments and was able to show in an ablation study that the inclusion of multi-step update in DQN yielded strong performance gains, but the study was limited, as it only compared single-step DQN and $n$-step DQN based on the same set of hyper-parameters. As single-step and $n$-step DQN are different algorithms, they may have a different set of optimal hyper-parameters. 
\citet{meng2021effect} showed empirically that a step size where $n = 1$ was comparatively more prone to inaccurate estimations of the q-value in comparison to values where $n > 1$. 
The authors also observed an improvement in training performance when the target was abstracted by an aggregate function over a set time horizon. A similar conclusion was again reached by \citet{chiang2020mixture} who experimented with a mixture of different step sizes and achieved stronger agent performance across a series of Atari test beds than single-step DQN. 
The authors hypothesised that the incremental improvement of agent performance was a result from an increased level of heterogeneity from the samples collected in the replay memory. While the study in general was limited by sample size having had only 3 random seeds as empirical evidence, it provided interesting insights into the potential of mixing step sizes to address both performance and estimation deficiencies experienced with single-step DQN. 
Other papers like \citep{deng2020value} also recognised the value of multi-step updates but their approach focused on discount rates that further regularises the reward accumulated to moderate the overestimation  potential in multi-step updates. 

Moreover, there also exists a wide body of work (\citep{van2016deep, hasselt2010double, anschel2017averaged, wang2016dueling, colas2018gep, dabney2018distributional, lan2019maxmin}) that while indirectly related to our investigation should be noted as they are widely used approaches to address divergent behaviours exhibited by DQN and serves as good benchmarks for our analysis. Of the large body of work, Double DQN and Average DQN will be used as benchmark algorithms in this paper,  both algorithms have shown efficacy with tempering overestimation  in DQN. Double DQN argues that the max operator used in DQN to both select and evaluate actions inadvertently leads the agent to be over optimistic in its value estimations. To address the problem of overestimation  induced the max operator, Double DQN decouples the use of the max operator in action and selection and evaluation. In contrast, Average DQN was also able to achieve a much more stable agent by taking the average of $k$ previous state action value estimates. This allowed the agent to produce more conservative and moderate state action values over time, thus reducing the overestimation . It is also possible as explored by \citep{sabry2019reduction} that overestimation  can be addressed through altering the neural network structure. \citet{sabry2019reduction} through their empirical investigations showed that the use of dropout layers may help alleviate overestimation .

Another related area similar to our approach are algorithmic methods that use contrastive learning and representation learning to remove noise and extract low-dimensional features from the observational space \citep{laskin2020curl, misra2020kinematic, pathak2017curiosity, efroni2021provable}. A major difference between the constrastive and representation learning approaches applied in a reinforcement learning context is that algorithms such as CURL \citep{laskin2020curl} act as prepossessing techniques that extract features from high-dimensional observation spaces and projects it to a more enriched lower dimensional space. On the other hand our algorithm builds a mechanism into the DQN to extract intermediate feature outputs from feed-forward layers to be used as training observations for an unsupervised learning algorithm, this component while non-trivial, represents a small portion of the algorithm.

Across the literature, overestimation is typically measured in terms of how far the $Q$-value diverges from the realisable true value of $Q$. In \citep{fujimoto2018addressing}, the true value is calculated as the average discounted return when the agent follows the current policy. In contrast, \citet{hasselt2010double, van2016deep} measured overestimation based on how far it deviated from the realisable range of the true absolute $Q$ value ($|Q|$) - this was made possible because the rewards in the environment was bounded between [-1, 1] and the gamma used was 0.99.

\section{Elastic Step Deep Q-Networks} \label{sec:esdqn}

The objective of Elastic Step DQN is to incorporate the ideas from Coarse Q-learning and multi-step DQN to reduce overestimation and improve the overall performance of DQN. To achieve this there needs to be a method of identifying whether two states are similar or dissimilar that is agnostic to state input structures or data type. One approach of doing this is through unsupervised clustering algorithms. Unsupervised clustering algorithms \cite{karoly2018unsupervised} are a class of algorithms designed to identify patterns in data without labels -- in this paper we use an algorithm known as HDBSCAN \cite{mcinnes2017hdbscan, mcinnes2017accelerated, hinneburg1999optimal} which has been shown to be very effective at identifying non-even transductive clusters. HDBSCAN is a clustering algorithm that extends on DBSCAN and was selected due to its robustness towards parameter selection, the algorithm performs well out of the box and required no parameter tuning of its own during the experiments. The algorithm much like its base variant, DBSCAN is also robust to the existence of outliers. KMeans and spectral clustering were also considered however both required the user to pre-determine the number of clusters which is unknown in the environments we experimented on \citep{hamerly2003learning, von2007tutorial, ng2001spectral}. The lack of hyper-parameter tuning naturally lent itself well to our experiments which needed to be transferable across different environments. 

There are two main components to this extension of DQN -- the first component is state information extraction and storage, here the aim is to find a suitable representation of the state to be used in the clustering algorithm. The second component involves the clustering and how its outputs are used to select the step horizon, which includes the reward accumulation.  

\begin{figure}
  \centering
  \includegraphics[width=1\textwidth]{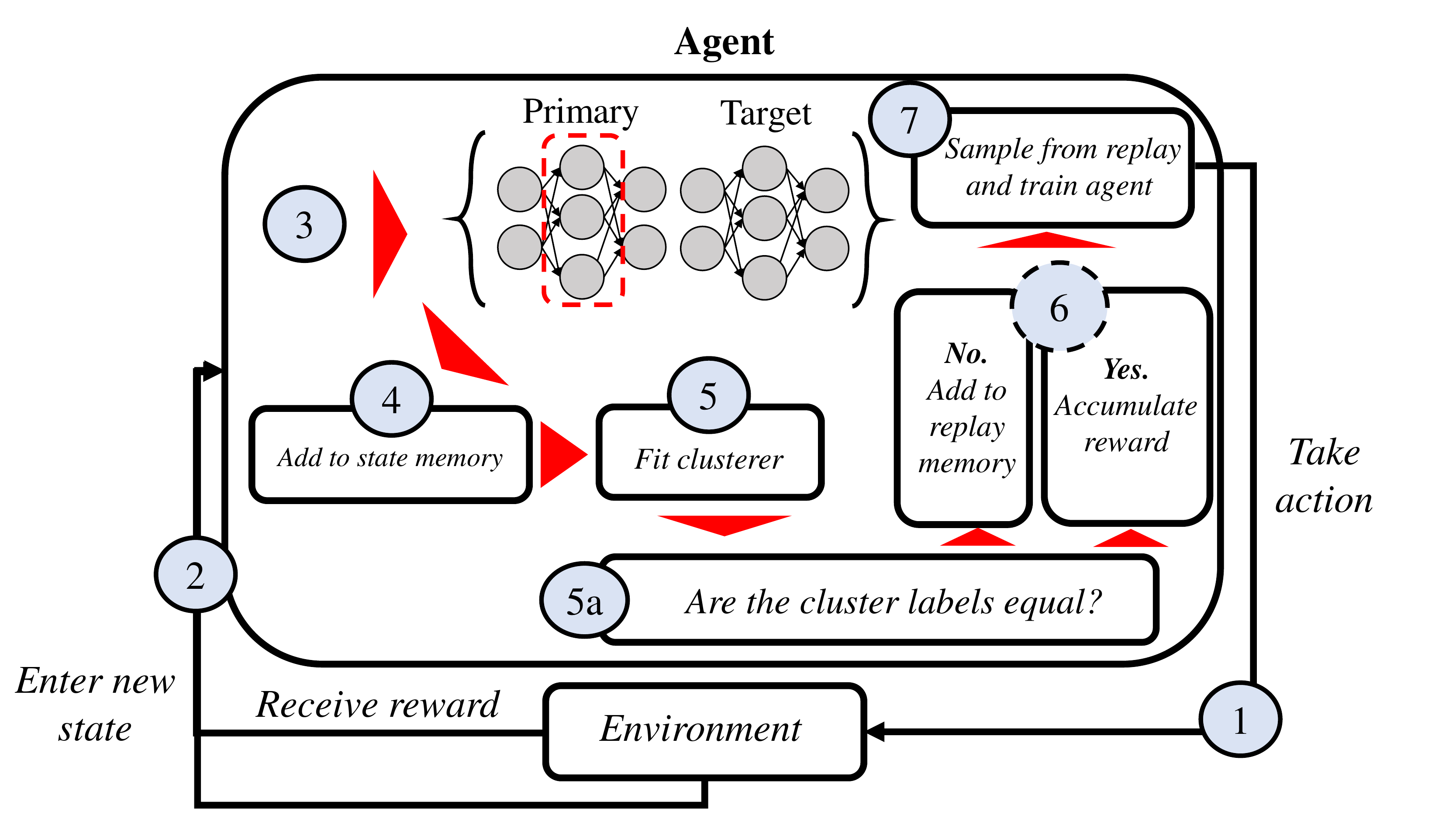}
  \caption{(1) the DQN agent observes $s_{t}$ and takes $a_{t}$ (2) receives $s_{t+1}$ and $r$ (3) output from $Q_{h}(s_{t+1})$ is stored in the state memory (4) a dataset is sampled from the state memory, (5) this is normalised  to unit variance and the clustering algorithm is fitted to the dataset (5a) cluster labels between $s_{t}$ and $s_{t+1}$ are compared. (6) If they are different then the tuple $(s_{t}, a_{t}, s_{t+1}, r)$ is added to the replay memory, if the cluster labels are not different, then the tuple is not added to the replay, the reward is accumulated. (7) Agent samples from replay and the primary network is trained. }
\label{fig:algo}
\end{figure}

\subsection{State information extraction and storage}

\subsubsection{State information extraction}

\begin{figure}[H]
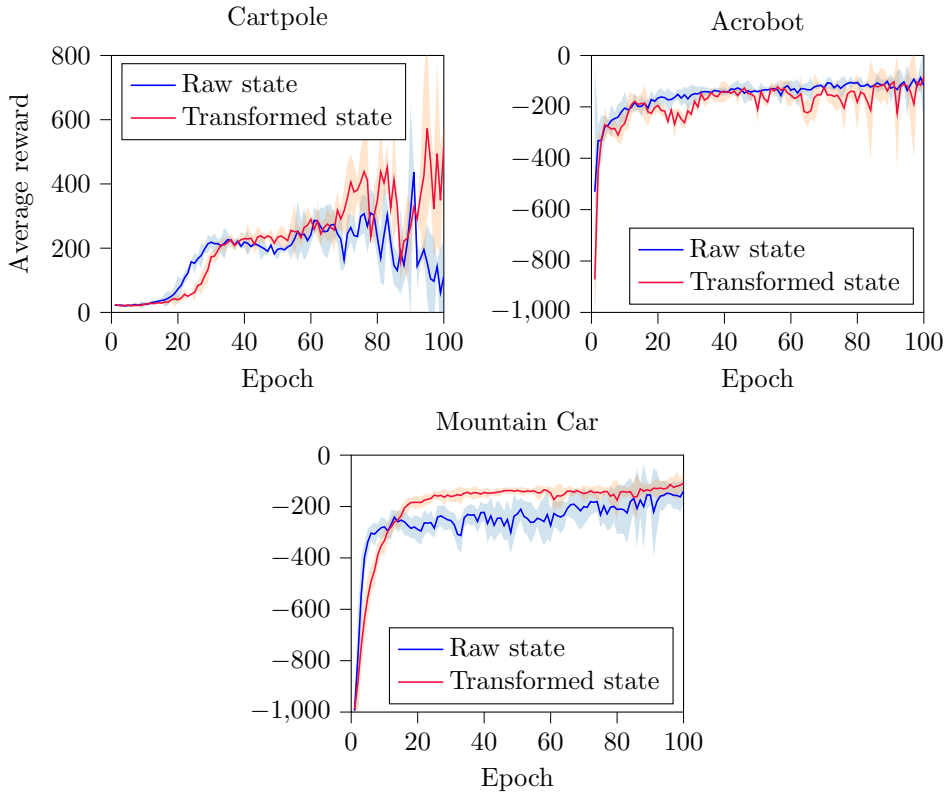

\centering
  \subfloat{\input{raw_state_perf}\label{fig:raw_state_perf_comp_cart}}
  \subfloat{\input{raw_state_perf_acro}\label{fig:raw_state_perf_comp_acro}}\hspace{1em}
  \subfloat{\input{raw_state_mc}\label{fig:raw_state_perf_comp_mc}}
  \caption{The blue line represents the training performance of Elastic Step DQN when the raw state is used while the red line represents the training performance when $Q(h)$ is used  as input into the clustering algorithm. The training performance is averaged over 30 seeds, and the shaded regioe n represents the 95 percent confidence interval.}\label{fig:raw_cat}
\end{figure}

A core component of Elastic Step DQN is the use of HDBSCAN to assign cluster labels to states. Previous research has indicated that \cite{talavera1999feature} high quality features generally plays a large role in the quality of clusters produced. Consequently we considered extracting the hidden node outputs as feature inputs into the unsupervised algorithm as previous research by \cite{ccayir2018feature} have indicated that it could be effective method to improve classifier performance. We experimented between an Elastic Step DQN variant that used raw state values as input against one that used outputs from the hidden node of the primary network and discovered that in both Cartpole and Mountain Car, we achieved much stronger training performance using the hidden node outputs as shown in Figure \ref{fig:raw_cat}. The performance difference between the use of raw state values vs hidden node outputs $Q_{h}(S)$ led us to choose the hidden node outputs as feature inputs into HDBSCAN.  

\subsubsection{State information storage}

Prior to the main algorithm loop a time-step counter $d$ is initialised at $d = 0$, the agent will observe the current state $s_{t}$ and take an action $a_{t}$ (label (1) in Figure \ref{fig:algo}). The agent will receive $s_{t+1}$ and $r_{t}$ (label (2) in Figure \ref{fig:algo}) and parse both $s_{t}$ and $s_{t+1}$ through the primary DQN network through to the hidden layer $Q_{h}$. The hidden layer output is extracted (label (3)) and stored in a state memory bank $B$ (label (4) in Figure \ref{fig:algo}). The state memory bank $B$ is populated with the outputs from the hidden node $Q_{h}(s_{t})$ generated from a uniform policy before any update has been applied. This early component is to ensure there is enough data to train the clustering algorithm in the initial stages. The state memory bank $B$ has a capacity of $H$, as new transformed states are added to the state memory bank, the older states are gradually replaced over time, this is to encourage diversity of more recent samples in the state memory bank. '


\subsection{State clustering}

Once the state memory bank $B$ is filled, the agent samples a dataset with replacement from $B$ -- each feature from this data-set is then standardised by removing the mean and scaled to unit variance. The data is sampled with replacement to ensure that the selected batch used to train the unsupervised algorithm is representative. Like most unsupervised clustering algorithm, HDBSCAN is less effective when the dataset is high dimensional \cite{mcinnes2017hdbscan, mcinnes2017accelerated, hinneburg1999optimal}, as a result if the number of features is greater than 30, then principal component analysis \cite{ding2004k} is applied to the dataset and only the top 30 components are taken. HDBSCAN is then trained on the dataset. Cluster labels are then assigned to $Q_{h}(s_t)$ and $Q_{h}(s_{t+1})$ (there is no limit set to the number of labels assigned to the particular dataset), if the cluster labels between $Q_{h}(s_t)$ and $Q_{h}(s_{t+1})$ are the same (i.e., they are similar states), then the reward and $\gamma$ are accumulated $\sum\gamma^{d}r_{t+d+1}$ and the time step counter $d$ is also incremented. 

If the cluster labels for $Q_{h}(s_{t})$ and $Q_{h}(s_{t+d+1})$ are different or either of them represents the terminal state, then the tuple ($s_{t}$, $a_{t}$, $r_{t+d+1}$, $s_{t+d+1}$,  $\gamma^{d}$) is then stored in the replay memory. In the instance where HDBSCAN detects outliers, each outlier is treated as a separate label, which means if both $s_t$ and $s_{t+d+1}$ are assigned an outlier label they are considered dissimilar. The performance of the algorithm is highly dependent on the quality of clusters generated, as a result it is important to retrieve a heterogeneous but representative sample of the environment for training. While HDBSCAN was the unsupervised clustering algorithm of choice for this paper, in theory any algorithm can be used as long as it generalises well to the environment.

{\centering
\begin{minipage}{1\linewidth}
\begin{algorithm}[H]

\caption{Elastic Step Deep Q-Network: $u$, $M$ and $T$ are all constants defined at the beginning of the algorithm, while $Q_{h}$ denotes the hidden layer of $Q$}\label{alg:cap}
\begin{algorithmic}[1]
\State Initialize a step counter ${d = 0}$, sample value $u$, max of episodes $M$, max of time-steps $T$, and target update interval

\State Initialize primary network $Q$ with weight $\theta$, target network $\hat Q$ with weight $\theta^{-} = \theta$ and replay buffer $D$ with capacity $N$
\State Initialize state memory bank $B$ with capacity $H$ and clustering algorithm $Clusterer$
\State Pre-fill $D$ with random experiences and $B$ with the hidden layer outputs of the same experiences
\For{episode = 1 to M}
    \State Reset the environment and observe the initial states
    \For{each time step to T}
        \State Select action $a_t$  using $\epsilon-greedy$
        \State Execute action in the environment and 
        \State observe the next state $s_{t+d+1}$ and reward $r_{t}$
        \State Pass $s_{t}$ and $s_{t+d+1}$ into $Q_{h}$
        \State Extract output and store in $B$ 
        \State With sample value $u$, randomly sample from $B$ 
        \State Train $Clusterer$ and assign cluster labels 
        \State to $Q_{h}(s_{t}$) and $Q_{h}(s_{t+d+1})$

        \If{cluster labels are not equal} 
            \State Store transition ($s_{t}$, $a_{t}$, $R_{t}$, $s_{t+d+1}$,  $\gamma^{d}$) in $D$
            \State Reset counter step counter ${d = 0}$
        \Else
            \State Increment step counter $d$
            \State $R_{t}+=\gamma^{d}$ $r_{t+d+1}$
        \EndIf 
        \State Randomly sample mini-batch from $D$
        \[y_{t+d+1} = \begin{cases} 
          R_t & \textit{for terminal $s_{t+d+1}$} \\
          R_t + \gamma^{d} max\hat Q(s_{t+d+1}, a_t) & \textit{for non terminal $s_{t+d+1}$} \\
       \end{cases}\]                                                                                                                  
    
    \State Update $\theta$ by gradient descent
    \State Update target network at each 
    \State target update interval
    \EndFor
\EndFor
\end{algorithmic}
\end{algorithm}
\end{minipage}
\par
}

\section{Experiments}

\subsection{Experimental setup} \label{sec:extset}

To test the efficacy of Elastic Step DQN, a large number of experiments were run across three separate OpenAI gym environments (Cartpole, Mountain Car and Acrobot) \citep{brockman2016openai}. For consistency, this paper utilises the  MushroomRL framework \citep{d2021mushroomrl} to implement all the algorithm. The use of smaller environments allows for a larger sample size of experiments to be run in comparison to what may be possible with larger environments \citep{ceron2021revisiting}. For each gym environment, eight seperate DQN implementations (single-step DQN, $2/3/4/6/8$ step DQN, Average DQN, Double DQN and Elastic Step DQN) were trained across 30 seeds for a range of different hyper-parameter settings. The best performing hyper-parameters across the 30 seeds were selected for comparison.  Each algorithm was trained on Cartpole, Acrobot, Mountain Car for respectively 40000, 40000 and 300000 time steps. The performance is then evaluated across 100 epochs. Each epoch represents $(time steps)/100$ (i.e. 400, 400 and 3000 time steps is a single epoch for Cartpole, Acrobot, Mountain Car for respectively) - the reward was captured both at the episodic and epoch level (with standard deviation, median and mean captured at each epoch) while the $|Q|$-values were captured at every 1000 time-step. Across each environment, if the termination condition is not met, the episode will terminate at the 1000th time step. 

From a hyper-parameter standpoint the search space for each algorithm was too large to be searched exhaustively, as a results only sensitive parameters were tuned while other parameters were based off of successful implementations of similar sized environments in the MushroomRL official repository. All agents received equal levels of hyper-parameter tuning, and each hyper-parameter set was trained for a full 30 seeds to check for statistical  significance -- as a result the best hyper-parameters for each algorithm in each environment is different (the hyper-parameters for each algorithm across the test-bed environments can be found in the Appendix A).

Each algorithm was evaluated on both the average final training reward and the average $|Q|$ achieved. A two sample t-statistic is calculated on the final training reward (as defined as the last 5 epochs) between the benchmark algorithm and Elastic Step DQN. To measure for divergent behaviour, this paper will refer to the true value of $|Q|$ as shown by \citep{thrun1993issues, van2018deep, hasselt2010double} where because the reward is bounded between [-1, 1] across all three environments and the $\gamma$ is 0.99, the true value of $|Q|$ is bounded by $\frac{1}{1-\gamma} = \frac{1}{1-0.99} = 100$. $|Q|$-values that exceed the upper bound of 100 are indicative of overestimation.  

Traditionally, the default upper bound of time-steps before the termination condition for Cartpole, Mountain Car and Acrobot is set at 200 time-steps. However, the upper bound artificially caps the amount of reward that the agent can receive which can obscure how well or poorly an agent can truly perform, consequently the termination condition was set at 1000 steps for all three environments.

\begin{figure*}[htp] 
\begin{subfigure}{0.49\textwidth} 
\centering     
\input{Cartpole_nstep.tex}
\caption{Cartpole training curve} \label{fig:nstep-a}
\end{subfigure}\hfill
\begin{subfigure}{0.49\textwidth} 
\centering     
\begin{tikzpicture}

\definecolor{cadetblue85154170}{RGB}{85,154,170}
\definecolor{crimson2411462}{RGB}{241,14,62}
\definecolor{darkgray176}{RGB}{176,176,176}
\definecolor{darkmagenta11540160}{RGB}{115,40,160}
\definecolor{gray143112113}{RGB}{143,112,113}
\definecolor{orangered2551010}{RGB}{255,101,0}
\definecolor{peru17214883}{RGB}{172,148,83}

\begin{axis}[
height=6cm,
legend cell align={left},
legend columns=2,
legend style={fill opacity=1, draw opacity=1, text opacity=1},
log basis x={10},
tick align=outside,
tick pos=left,
width=7cm,
x grid style={darkgray176},
xlabel={Average absolute Q-value (log)},
xmin=20.2482899049114, xmax=44143.2201539436,
xmode=log,
xtick style={color=black},
xtick={1,10,100,1000,10000,100000,1000000},
xticklabels={
  \(\displaystyle {10^{0}}\),
  \(\displaystyle {10^{1}}\),
  \(\displaystyle {10^{2}}\),
  \(\displaystyle {10^{3}}\),
  \(\displaystyle {10^{4}}\),
  \(\displaystyle {10^{5}}\),
  \(\displaystyle {10^{6}}\)
},
y grid style={darkgray176},
ylabel={Average reward},
ymin=-0.912600166996924, ymax=249.476563726773,
ytick style={color=black}
]
\addplot [draw=cadetblue85154170, fill=cadetblue85154170, mark=*, only marks, opacity=0.5]
table{%
x  y
33.2844306893304 78.126953125
30.5377795780988 82.9896265560166
32.1961070037022 84.568710359408
31.5189326026201 92.8097447795824
31.7859645500742 100.0025
33.2738891886123 107.529569892473
33.104716665595 64.0016
33.3407301913612 79.6832669322709
36.0111717954263 95.9256594724221
36.5896327079885 86.0236559139785
32.4343544308119 93.6791569086651
31.1003834862262 92.168202764977
32.2485945507329 76.192380952381
32.731415196766 104.169270833333
33.346257190406 104.441253263708
34.5170243951522 82.646694214876
33.4441737344466 81.1379310344828
29.9633159615267 86.2090517241379
33.8290159965541 113.317280453258
33.7678437628575 101.012626262626
31.2228046573907 85.6552462526767
33.0573386522874 81.3028455284553
32.6655642089896 85.6552462526767
31.8115774201855 96.6207729468599
34.1361622746423 62.2099533437014
32.7023062888287 120.123123123123
32.3445027362697 95.0142517814727
33.5319214109172 197.049261083744
44.1141386835877 81.969262295082
33.7305628448024 99.0123762376238
};
\addlegendentry{2S DQN}
\addplot [draw=darkmagenta11540160, fill=darkmagenta11540160, mark=*, only marks, opacity=0.5]
table{%
x  y
45.4930811971981 58.825
44.1770830573246 195.126829268293
45.5673158687611 66.226821192053
37.4402923122711 101.52538071066
40.3381799472436 102.830334190231
43.5319453037616 153.85
40.870019819124 169.495762711864
43.5001599829026 119.050595238095
44.8440184638701 102.043367346939
41.3909344048724 127.391719745223
52.6442076892793 136.522184300341
42.0357458648294 98.7679012345679
42.2632476952504 83.68410041841
38.955699372419 66.0082508250825
41.244583265356 82.646694214876
66.8499450262446 70.6731448763251
47.8084665874377 118.697329376855
40.4537543339545 111.113888888889
41.6923299514152 104.441253263708
41.4651465956725 82.9896265560166
42.9527925694939 129.873376623377
44.2973469569616 68.0289115646258
39.9866289253116 170.217021276596
44.317160342177 156.25390625
48.0131768509276 98.5246305418719
43.2144800483137 100.0025
40.8290471491858 152.675572519084
40.3071105660632 76.925
43.8556202671006 108.40379403794
40.3179815079562 81.969262295082
};
\addlegendentry{3S DQN}
\addplot [draw=peru17214883, fill=peru17214883, mark=*, only marks, opacity=0.5]
table{%
x  y
43.14504094318 217.396739130435
41.4039758015171 59.9715142428786
70.9361674201116 77.6718446601942
44.1475945185646 75.9032258064516
43.7122461672164 75.9032258064516
50.6035450915061 88.1079295154185
42.3324069439687 221
42.3118467432581 156.25390625
131.51168422794 57.1442857142857
39.7059020437889 66.779632721202
52.3061285040312 25.2691092861655
41.4371582322329 63.4936507936508
79.6428677174386 40.4050505050505
153.934530332282 52.4259501965924
35.6036962945014 51.0868454661558
50.312970530178 103.361757105943
40.0182061745055 153.260536398467
51.8615488132615 73.8025830258303
68.4048637485988 105.543535620053
54.6499945784494 47.06
51.6142627321482 70.4242957746479
40.4263849424042 103.629533678756
43.4979564034317 91.3264840182648
45.3998902539782 46.5669383003492
37.5370742439575 75.1898496240601
41.2564386352077 76.192380952381
37.6004820679195 93.0255813953488
33.1571679273672 70.0542907180385
49.6599497190177 57.3080229226361
121.240537844734 36.4307832422586
};
\addlegendentry{4S DQN}
\addplot [draw=gray143112113, fill=gray143112113, mark=*, only marks, opacity=0.5]
table{%
x  y
3563.81995242877 11.8451288125555
55.5958809752271 59.9715142428786
265.072249587615 24.7224969097651
34.0597716540366 101.26835443038
40.3617394777954 51.3491655969191
47.4657531680018 93.4602803738318
42.4466134129815 66.6683333333333
104.28935936237 44.9955005624297
132.688073252009 66.8913043478261
28.7168793710053 79.209900990099
549.488604836177 22.4220852017937
33.1040233311534 65.3611111111111
42.5579843445245 56.3394366197183
170.099279473416 45.6632420091324
53.481760894607 43.7169398907104
41.6562970717844 126.987301587302
56.783087525475 49.5061881188119
41.0166018311709 43.7647702407002
5455.26993723196 19.1026743075454
76.644762539611 30.9605263157895
28.8684637262538 42.1506849315069
64.9587292672709 123.842105263158
63.8728462066159 32.1810136765889
46.2384701198749 96.3879518072289
408.397349761847 82.1375770020534
56.6230107257813 33.0041254125413
91.2066503739089 30.9605263157895
31.0332045059562 68.7302405498282
93.4730067230895 43.0118279569893
737.435072862525 23.2428820453225
};
\addlegendentry{6S DQN}
\addplot [draw=orangered2551010, fill=orangered2551010, mark=*, only marks, opacity=0.5]
table{%
x  y
668.165851867646 12.7594896331738
815.796504798853 18.1
65.0417452653881 24.3908536585366
1725.57422730614 17.1899441340782
41.7355939474687 49.8144458281445
35.1391311179541 51.2833333333333
44.490244572537 77.8229571984436
2536.94757722337 17.9779775280899
36.9006016225355 32.922633744856
101.146220865603 16.6117109634551
649.729936300016 39.2166666666667
823.304727302285 19.0480952380952
54.4504763072528 45.3012457531144
378.400393487355 18.4591601292109
178.319501156505 111.423398328691
83.6813293304771 20.0305458187281
31125.4126002232 10.4687254645381
2699.65781452776 16.4342645850452
7786.01024740604 14.8206743238236
939.489460017144 16.7228260869565
45.2771125784829 36.2985480943739
44.1383012154818 43.5740740740741
1471.74188320488 31.6213438735178
68.5966248135835 32.1293172690763
53.3843888667956 42.7361111111111
30332.482165749 18.7533989685888
129.196138846022 24.3315085158151
45.1704156016126 46.9495305164319
4180.72568963631 14.2656918687589
427.528755050967 38.3518696069032
};
\addlegendentry{8S DQN}
\addplot [draw=black, fill=black, mark=*, only marks, opacity=0.5]
table{%
x  y
51.8447982086455 150.37969924812
52.0791065372124 135.138513513514
52.113207612893 145.458181818182
54.1333718683578 99.258064516129
53.3576080669716 139.376306620209
52.1947536908954 134.683501683502
52.3079692293651 151.518939393939
52.9373104477778 155.042635658915
52.0817992715239 162.605691056911
52.2342164899632 133.782608695652
52.2946379200537 166.670833333333
52.2564731393628 144.93115942029
52.1997107432664 161.947368421053
52.5293874831215 135.138513513514
51.7943861291824 190.480952380952
51.6654264584932 169.495762711864
52.7640217154294 132.016501650165
53.125697607556 148.702602230483
52.3835986580502 149.816479400749
52.460532563898 132.453642384106
52.6193761600412 139.863636363636
52.4744847605295 158.734126984127
53.7765719955891 148.702602230483
52.4151032639258 152.095057034221
52.0365373900548 140.354385964912
53.2360363316819 140.354385964912
52.4999887265235 148.151851851852
54.5055583420977 158.734126984127
53.4968275308911 117.304985337243
51.8347669138506 150.947169811321
};
\addlegendentry{DQN}
\addplot [draw=crimson2411462, fill=crimson2411462, mark=*, only marks, opacity=0.5]
table{%
x  y
53.1382784269443 163.934426229508
54.801753745042 144.927536231884
51.745071349404 152.091254752852
52.2429893382573 191.387559808612
52.4788034739107 182.648401826484
52.322190544048 185.185185185185
51.639544120086 173.913043478261
52.3506146260925 173.160173160173
53.5757508374564 160.642570281125
52.0159820921732 182.648401826484
51.9739450401165 212.765957446808
52.6028309701107 144.404332129964
52.012171808641 215.05376344086
52.3961330478277 183.48623853211
52.7361803044658 185.185185185185
52.4929489501115 176.211453744493
51.3248255539317 182.648401826484
52.1216617645323 166.666666666667
52.2354855247222 193.236714975845
52.3209972846709 184.331797235023
52.4837400361046 147.60147601476
52.0540810328871 238.095238095238
52.9053567497663 165.97510373444
52.5289769218884 176.211453744493
52.0977092141166 173.160173160173
51.906208710302 205.128205128205
52.9819895374238 160.642570281125
52.277717189493 176.991150442478
51.7467916965526 173.913043478261
52.299540485207 186.046511627907
};
\addlegendentry{ES DQN}
\addplot [semithick, black, dashed, forget plot]
table {%
100 -0.912600166996928
100 249.476563726773
};
\end{axis}

\end{tikzpicture}
\caption{Cartpole reward vs. estimation analysis} \label{fig:nstep-b}
\end{subfigure}

\begin{subfigure}{0.49\textwidth} 
\centering     
\input{Acrobot_nstep.tex}
\caption{Acrobot training curve} \label{fig:nstep-c}
\end{subfigure}\hfill
\begin{subfigure}{0.49\textwidth} 
\centering     
\begin{tikzpicture}

\definecolor{cadetblue85154170}{RGB}{85,154,170}
\definecolor{crimson2411462}{RGB}{241,14,62}
\definecolor{darkgray176}{RGB}{176,176,176}
\definecolor{darkmagenta11540160}{RGB}{115,40,160}
\definecolor{gray143112113}{RGB}{143,112,113}
\definecolor{orangered2551010}{RGB}{255,101,0}
\definecolor{peru17214883}{RGB}{172,148,83}

\begin{axis}[
height=6cm,
tick align=outside,
tick pos=left,
width=7cm,
x grid style={darkgray176},
xlabel={Average $|Q|$-value},
xmin=0, xmax=100,
xtick style={color=black},
y grid style={darkgray176},
ylabel={Average reward},
ymin=-971.380469480641, ymax=-65.1031641623543,
ytick style={color=black}
]
\addplot [draw=cadetblue85154170, fill=cadetblue85154170, mark=*, only marks, opacity=0.5]
table{%
x  y
67.7821908680633 -253.929936305732
67.5918848027363 -149.458646616541
66.5382515913375 -185.153488372093
68.172371723558 -314.181102362205
63.3057712686062 -160.383064516129
68.9880316552624 -143.494584837545
66.8842266646251 -262.335526315789
65.5554503057145 -122.907120743034
63.5180771323264 -147.7843866171
65.5870133775502 -146.686346863469
66.4586152410388 -237.25
70.4681218092576 -499.375
60.8010320198238 -280.87323943662
62.0869044927075 -113.022792022792
66.7780167091899 -124.457680250784
63.3115269217238 -137.96875
68.8572976314485 -115.685131195335
65.7042436408095 -121.018292682927
62.7086833514713 -110.793296089385
64.0864306714483 -130.654605263158
71.1829691424131 -438.923076923077
61.3513921071425 -137.487889273356
70.9158867247462 -241.581818181818
55.4681555611052 -119.186186186186
51.1908012511194 -190.497607655502
53.68970016528 -106.297587131367
69.0837871217593 -269.452702702703
67.0497807582349 -159.730923694779
68.3439396191955 -195.191176470588
66.6579348598458 -141.423487544484
};
\addplot [draw=darkmagenta11540160, fill=darkmagenta11540160, mark=*, only marks, opacity=0.5]
table{%
x  y
38.4758341147006 -403.333333333333
39.752300853698 -469.976470588235
39.9813380807504 -332.55
39.1589058930427 -740.407407407407
39.2367782219797 -347.034782608696
39.1525317661166 -309.333333333333
39.0662183393002 -321.790322580645
39.1712968167722 -139.926056338028
39.1581723944068 -178.470852017937
39.5900892932057 -677.576271186441
39.2244619231939 -252.310126582278
38.8136896877497 -359.621621621622
39.869385766007 -198.139303482587
39.7085558177352 -286.935251798561
40.2977360124886 -221.372222222222
39.4594679415554 -299.932330827068
39.9546776486874 -624.5625
39.5359923822731 -247.60248447205
39.8421909819588 -218.901098901099
40.1202772074267 -211.898936170213
39.3145144753456 -264.066225165563
39.7998946430534 -146.686346863469
22.9279348884046 -415.958333333333
40.1087435446516 -299.977443609023
39.6929624504864 -276.9375
38.8838798582375 -128.103225806452
39.4303418230414 -280.859154929577
39.437175000307 -359.630630630631
39.2643755548596 -299.90977443609
39.3550700661182 -369.648148148148
};
\addplot [draw=peru17214883, fill=peru17214883, mark=*, only marks, opacity=0.5]
table{%
x  y
80.3711118524179 -250.729559748428
81.0302816754825 -122.144615384615
82.4131100738987 -547.397260273973
80.1955193416372 -286.956834532374
72.9179874424174 -133.308724832215
76.9486448566541 -176.088495575221
75.5075220447406 -909.022727272727
80.2186079863966 -165.066390041494
76.7482523644999 -126.057142857143
80.8698396111324 -140.921985815603
77.9242915534802 -424.925531914894
76.1626360555619 -179.279279279279
78.6032962862983 -165.0622406639
78.4656861450717 -246.067901234568
82.308135716103 -275.013793103448
80.242328465467 -438.89010989011
82.3589867096603 -134.664406779661
74.132095252198 -110.790502793296
84.3902007464826 -197.128712871287
79.9426788642377 -107.755434782609
82.0587047811657 -160.391129032258
78.1732351653501 -420.305263157895
82.7583368276656 -570.9
79.0196253018111 -180.095022624434
81.2571297127001 -160.383064516129
82.3704687655397 -124.460815047022
81.6917373920098 -213.042780748663
82.84798146106 -395.306930693069
81.7499335592389 -403.373737373737
83.7787947060414 -119.545180722892
};
\addplot [draw=gray143112113, fill=gray143112113, mark=*, only marks, opacity=0.5]
table{%
x  y
92.6817516095042 -415.947916666667
93.6962481612414 -226.414772727273
92.7968456511378 -341.076923076923
93.0304294542432 -262.302631578947
89.9906395169675 -206.367875647668
92.1300620543063 -223.842696629213
93.2849205191672 -726.963636363636
93.053480774954 -234.429411764706
91.9093527284443 -188.687203791469
94.0082246602207 -525.763157894737
93.1468536037564 -207.453125
93.6566060041666 -493.185185185185
87.1385531217277 -525.802631578947
93.1319376579225 -221.35
92.6525233858764 -434.065217391304
93.9668861165673 -387.631067961165
93.2923916683435 -353.194690265487
91.9438448616982 -165.758333333333
93.8752018952668 -799.78
92.7562987613976 -429.376344086022
91.1764383448005 -128.103225806452
92.5617476959825 -192.338164251208
92.1986901815295 -499.425
92.1182803535938 -193.281553398058
86.8416027656123 -150.030188679245
86.687301431115 -140.427561837456
93.0061454789832 -243.05487804878
91.0371377863556 -286.942446043165
92.7209623936907 -264.05298013245
93.2540094055444 -277
};
\addplot [draw=orangered2551010, fill=orangered2551010, mark=*, only marks, opacity=0.5]
table{%
x  y
97.8282665439785 -487.158536585366
97.2810083586097 -269.452702702703
97.1247326006442 -316.634920634921
97.2931555945098 -644.709677419355
98.1255166169599 -191.413461538462
97.5296545426786 -243.042682926829
97.6717653383613 -391.392156862745
92.2873123716116 -170.029914529915
97.0660592671484 -220.110497237569
97.8903633230716 -273.157534246575
97.387968770957 -459.091954022988
97.576864484188 -634.444444444444
97.777873523736 -596.522388059701
96.6177826135844 -332.575
97.5338080347985 -420.368421052632
97.499952457495 -267.604026845638
97.8888402910054 -314.141732283465
97.6864640547722 -306.861538461538
97.9038782151461 -235.816568047337
97.5735274845779 -518.935064935065
97.8317938576058 -525.710526315789
97.8753585739344 -930.186046511628
96.6130889115781 -225.107344632768
95.5522373031467 -178.470852017937
96.7313154072978 -173.013043478261
96.2783518283337 -186.014018691589
97.983414527908 -344.034482758621
97.8306425509974 -171.504310344828
97.5034167215437 -587.661764705882
97.9985086786769 -289.04347826087
};
\addplot [draw=crimson2411462, fill=crimson2411462, mark=*, only marks, opacity=0.5]
table{%
x  y
85.1360123744607 -151.751908396947
82.7828237935424 -127.684887459807
72.9742866926759 -114.664739884393
69.2152074534535 -108.049046321526
71.7642063218743 -117.762611275964
77.0063308478907 -117.05604719764
75.707817283231 -127.688102893891
90.1049461709581 -234.417647058824
79.8253484727412 -128.944805194805
88.8462948001295 -249.15
72.9956557542115 -130.648026315789
68.3662480779976 -109.555248618785
70.0287333279029 -109.861495844875
89.6844371589184 -260.607843137255
83.3945883918494 -136.058219178082
84.6927877131566 -143.494584837545
80.6348654646121 -159.730923694779
72.8721807329252 -116.360703812317
83.2234592260331 -141.935714285714
90.9847751316935 -434.086956521739
64.2679866075695 -109.861495844875
89.0938635350794 -356.392857142857
65.3365931466967 -112.050847457627
87.0988399514228 -244.564417177914
77.7897523649991 -119.909365558912
75.432720368509 -119.545180722892
78.9175407677323 -147.244444444444
68.2732234408811 -109.250688705234
87.3975154962659 -171.5
79.9345115423203 -150.598484848485
};
\addplot [draw=black, fill=black, mark=*, only marks, opacity=0.5]
table{%
x  y
65.7107016790092 -130.648026315789
68.0037603492558 -130.219672131148
68.2440700667441 -159.088
69.9076334176168 -175.303964757709
67.5473768457875 -130.648026315789
68.081295200789 -144.007246376812
67.4157192988098 -132.852842809365
68.1185433421984 -159.730923694779
65.0583470086798 -123.291925465839
64.1444124869891 -126.458598726115
67.0094834406406 -132.406666666667
68.0218284050718 -131.082508250825
67.3944819811627 -138.933566433566
68.2608020928979 -182.596330275229
65.6805821869701 -126.053968253968
68.5302748423256 -192.347826086957
69.3075773464039 -208.539267015707
67.6009304096505 -157.81746031746
66.7050589160919 -127.688102893891
67.085341365321 -167.869198312236
64.475303810589 -116.023391812865
68.7815392960534 -160.379032258065
66.8009513627961 -139.926056338028
69.0301169711694 -178.470852017937
69.7406805001445 -163.02868852459
65.9507326008365 -152.930769230769
68.2914370252565 -167.163865546218
67.1398611397028 -128.521035598706
68.8910564876229 -173.008695652174
66.1106191979662 -131.963455149502
};
\addplot [semithick, black, dashed]
table {%
100 -971.380469480641
100 -65.1031641623543
};
\end{axis}

\end{tikzpicture}
\caption{Acrobot reward vs. estimation analysis} \label{fig:nstep-d}
\end{subfigure}

\begin{subfigure}{0.49\textwidth} 
\centering     
\input{mc_nstep}
\caption{Mountain Car training curve} 
\label{fig:nstep-e}
\end{subfigure}\hfill
\begin{subfigure}{0.49\textwidth} 
\centering     
\begin{tikzpicture}

\definecolor{cadetblue85154170}{RGB}{85,154,170}
\definecolor{crimson2411462}{RGB}{241,14,62}
\definecolor{darkgray176}{RGB}{176,176,176}
\definecolor{darkmagenta11540160}{RGB}{115,40,160}
\definecolor{gray143112113}{RGB}{143,112,113}
\definecolor{orangered2551010}{RGB}{255,101,0}
\definecolor{peru17214883}{RGB}{172,148,83}

\begin{axis}[
height=6cm,
log basis x={10},
tick align=outside,
tick pos=left,
width=7cm,
x grid style={darkgray176},
xlabel={Average $|Q|$-value},
xmin=50.9604756063092, xmax=21809.2710083462,
xmode=log,
xtick style={color=black},
xtick={1,10,100,1000,10000,100000,1000000},
xticklabels={
  \(\displaystyle {10^{0}}\),
  \(\displaystyle {10^{1}}\),
  \(\displaystyle {10^{2}}\),
  \(\displaystyle {10^{3}}\),
  \(\displaystyle {10^{4}}\),
  \(\displaystyle {10^{5}}\),
  \(\displaystyle {10^{6}}\)
},
y grid style={darkgray176},
ylabel={Average reward},
ymin=-1042.51847005988, ymax=-107.185462075848,
ytick style={color=black}
]
\addplot [draw=cadetblue85154170, fill=cadetblue85154170, mark=*, only marks, opacity=0.5]
table{%
x  y
97.1197014939007 -819.674863387978
98.6581987628287 -983.609836065574
98.1712529944317 -931.680124223602
98.5017957936832 -1000.00333333333
97.8784877268049 -862.07183908046
96.1140725386673 -1000.00333333333
98.4753340455493 -993.380794701987
96.6281119344778 -974.029220779221
98.6932457114555 -700.93691588785
98.5749479210364 -524.477272727273
98.3619749759428 -612.24693877551
98.067847214515 -1000.00333333333
98.749708227788 -990.102310231023
98.6063882995854 -465.117829457364
98.7138797866983 -1000.00333333333
98.7293970107849 -983.609836065574
98.0014476547711 -366.749388753056
97.1543357580912 -1000.00333333333
98.2671154346315 -1000.00333333333
98.8116302489072 -990.102310231023
98.5218249466437 -983.609836065574
97.8425238296249 -993.380794701987
98.7999319230773 -1000.00333333333
98.5023674341481 -996.681063122924
98.2996248420937 -996.681063122924
98.6273842838216 -961.541666666667
96.653370627623 -719.426858513189
98.1755201841754 -396.82671957672
98.2321731738682 -282.2210724365
98.4833169948134 -1000.00333333333
};
\addplot [draw=darkmagenta11540160, fill=darkmagenta11540160, mark=*, only marks, opacity=0.5]
table{%
x  y
98.2505361415446 -819.674863387978
98.9616305155626 -983.609836065574
98.1730470988083 -931.680124223602
99.1084481375665 -1000.00333333333
98.2060687807613 -862.07183908046
97.9951509608606 -1000.00333333333
98.9175803567695 -993.380794701987
98.4881565752279 -974.029220779221
98.7957182593369 -700.93691588785
98.9219222481961 -524.477272727273
99.1729704928156 -612.24693877551
98.5924665480623 -1000.00333333333
98.9847744856874 -990.102310231023
98.6612509041096 -465.117829457364
98.3886535012817 -1000.00333333333
99.0800016943892 -983.609836065574
98.6415566185831 -366.749388753056
98.7351022767199 -1000.00333333333
98.7901923201979 -1000.00333333333
98.7538265500331 -990.102310231023
98.4641560832649 -983.609836065574
98.6210116614366 -993.380794701987
98.6361621084843 -1000.00333333333
98.2161996438097 -996.681063122924
98.3326899803348 -996.681063122924
98.8601056133862 -961.541666666667
98.1973496722674 -719.426858513189
98.4515866026907 -396.82671957672
98.9333966004654 -282.2210724365
98.4596318891487 -1000.00333333333
};
\addplot [draw=peru17214883, fill=peru17214883, mark=*, only marks, opacity=0.5]
table{%
x  y
98.1962019520352 -1000.00333333333
98.4513979364961 -1000.00333333333
98.6591362480298 -874.638483965015
98.8699876742896 -980.395424836601
98.4252548145419 -983.609836065574
97.9254272273461 -977.201954397394
98.8706851209972 -983.609836065574
97.9145628742019 -990.102310231023
98.2560068263522 -993.380794701987
98.4343469718319 -806.454301075269
98.4717681391261 -964.633440514469
98.4256801306 -882.355882352941
98.6651045295283 -604.840725806452
98.4088732620146 -486.22528363047
97.8938769069006 -990.102310231023
98.7963726691729 -990.102310231023
97.5656356685355 -993.380794701987
97.5083748929504 -678.735294117647
98.3787593768525 -432.901875901876
98.6846858706725 -748.13216957606
97.0994260977987 -961.541666666667
98.2718284668638 -980.395424836601
98.5017794084926 -996.681063122924
98.8619212132025 -964.633440514469
98.2061934843135 -990.102310231023
98.3407531216765 -427.961483594864
98.2585886541619 -911.857142857143
98.3316510420854 -601.204408817635
98.6964019178202 -1000.00333333333
98.0783505690755 -990.102310231023
};
\addplot [draw=gray143112113, fill=gray143112113, mark=*, only marks, opacity=0.5]
table{%
x  y
98.2659249006805 -955.417197452229
98.6312873481556 -996.681063122924
98.8608593089899 -1000.00333333333
98.6799893125209 -940.442006269593
98.6642979117572 -952.384126984127
98.5259111686864 -996.681063122924
98.4880216870482 -990.102310231023
98.4388084634979 -996.681063122924
98.1812639870039 -996.681063122924
98.6561924995031 -996.681063122924
98.8894401424977 -993.380794701987
98.7605642569959 -645.163440860215
98.6656002620168 -1000.00333333333
98.5743612583177 -996.681063122924
98.2153816656031 -996.681063122924
98.8278381469617 -996.681063122924
98.4648305150323 -1000.00333333333
98.4242908661034 -867.054913294798
98.4842068833425 -986.845394736842
98.598643265587 -1000.00333333333
98.3039824337104 -906.347432024169
98.0746876477928 -835.657381615599
98.46009202357 -983.609836065574
98.8639317671904 -986.845394736842
98.1423128260031 -996.681063122924
98.939924658193 -787.404199475066
98.1876178695508 -368.099386503067
98.6732274248618 -589.392927308448
98.449121800965 -1000.00333333333
98.4196275141866 -537.636200716846
};
\addplot [draw=orangered2551010, fill=orangered2551010, mark=*, only marks, opacity=0.5]
table{%
x  y
98.8898350398747 -619.836776859504
98.5904284468904 -1000.00333333333
98.4729911681862 -996.681063122924
98.6031471417646 -983.609836065574
98.7076781722449 -729.929440389294
98.8200538309697 -728.157766990291
98.5088980994294 -986.845394736842
98.4366584560462 -996.681063122924
98.7606896227771 -879.768328445748
98.3242747802709 -993.380794701987
98.6705751103258 -840.33893557423
98.9482965019625 -955.417197452229
98.7305293932024 -996.681063122924
99.063594276614 -705.884705882353
98.1890521394702 -1000.00333333333
98.8417222756038 -1000.00333333333
98.9074412751104 -1000.00333333333
98.1940598413964 -1000.00333333333
98.9475864201565 -748.13216957606
98.9414711410854 -993.380794701987
98.4324491232193 -990.102310231023
98.5423699958387 -993.380794701987
98.9237182191416 -970.877022653722
98.3274836957049 -895.525373134328
98.5836119736934 -1000.00333333333
98.8714983952415 -381.680661577608
98.8335459921756 -616.018480492813
98.4768184654856 -967.745161290323
98.8896848211936 -970.877022653722
98.7789156709434 -636.944798301486
};
\addplot [draw=black, fill=black, mark=*, only marks, opacity=0.5]
table{%
x  y
109.962662358181 -215.363244795406
16558.9086139251 -185.874225526642
75.1264300180001 -201.342953020134
859.641626400496 -193.741430700447
520.733161609965 -255.537478705281
1806.16350289864 -169.683823529412
1136.5880064261 -175.644613583138
587.456572343173 -217.392028985507
5295.38594887539 -173.712217718587
83.2526190541965 -236.780584056827
128.040569121769 -324.325405405405
7301.39632183378 -184.049693251534
227.784219126798 -242.915789473684
906.313691052888 -251.678691275168
95.9241994879023 -872.095930232558
538.868994449864 -258.176419965577
1431.83721247973 -186.567786069652
2134.20332315799 -251.890008396306
77.0058686223579 -242.71925566343
4546.40588786596 -184.616
84.6664374150422 -204.639154160982
123.104270154029 -330.034103410341
559.283498898714 -209.35170969993
4961.33984211051 -172.513513513514
431.577274285755 -207.326192121631
1448.46761177852 -224.383694839192
77.4362606189187 -208.768963117606
634.137507439096 -243.507305194805
90.7181992409742 -385.110397946085
93.7820611754572 -465.840062111801
};
\addplot [draw=crimson2411462, fill=crimson2411462, mark=*, only marks, opacity=0.5]
table{%
x  y
87.4032555854313 -325.732899022801
75.1948773181905 -187.5
70.7116194960881 -156.168662155128
93.4171620706616 -607.287449392713
93.2829274051342 -480.769230769231
78.4516761894558 -207.03933747412
75.6048347126083 -180.722891566265
93.6626190212353 -993.377483443709
80.0346159353772 -232.018561484919
67.1186035942963 -149.700598802395
76.7258400738423 -185.873605947955
74.5741074135739 -172.910662824207
72.8757526499278 -163.934426229508
82.8780591135395 -232.018561484919
76.3897203204122 -193.673337637185
76.3611757391264 -166.389351081531
81.7648031487853 -209.643605870021
93.4038225483601 -940.438871473354
102.536748023094 -225.225225225225
88.5638910839599 -271.24773960217
78.1925209714585 -192.926045016077
70.0702995542806 -157.068062827225
72.3311814797188 -165.562913907285
93.3047903841628 -625
74.3303458545993 -186.219739292365
81.9122678879437 -207.325501036628
80.9813911031501 -202.292650033715
81.1948163409422 -262.46719160105
91.6590520335263 -375
93.350749736845 -472.44094488189
};
\addplot [semithick, black, dashed]
table {%
100 -1042.51847005988
100 -107.185462075848
};
\end{axis}

\end{tikzpicture}
\caption{Mountain Car reward vs. estimation analysis} \label{fig:nstep-f}
\end{subfigure}
\caption{\ref{fig:nstep-a}, \ref{fig:nstep-c} and \ref{fig:nstep-e} compare the training curves between the $2/3/4/6/8$-Step DQN updates against Elastic Step DQN and single-step DQN. They show the average episodic rewards across the training epochs. The solid line represents the average reward while the shaded region represents the 95 percent confidence interval. The scatter plots \ref{fig:nstep-b}, \ref{fig:nstep-d} and \ref{fig:nstep-f} map the relationship between the average reward achieved across each experiment against the average $|Q|$-values from that experiment. The dotted dash line represents the upper bound of the true $|Q|$-value estimate.} \label{fig:nstep}
\end{figure*}

\begin{figure*}[htp]
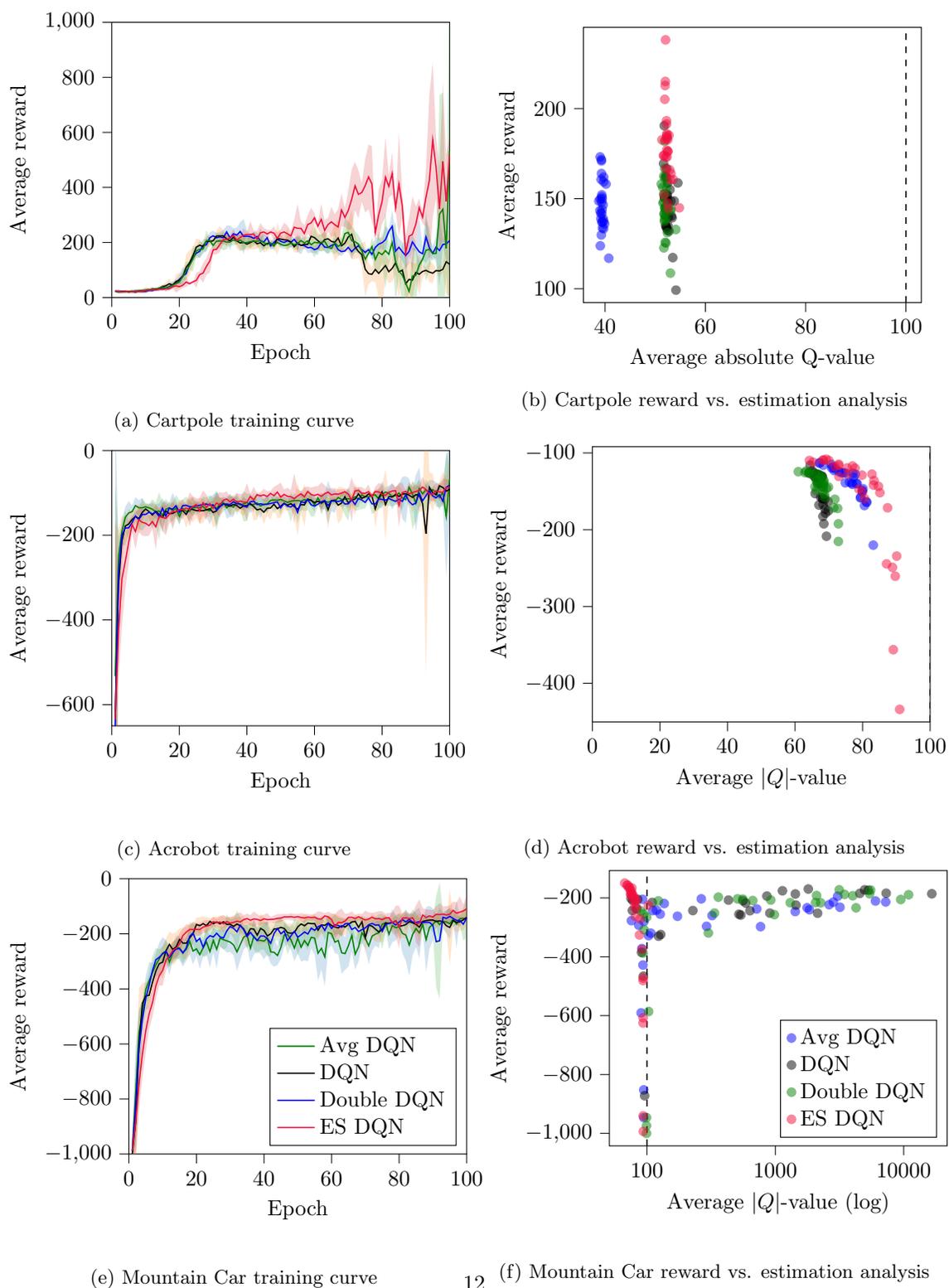
 
\begin{subfigure}{0.49\textwidth}
\centering     
\input{Cartpole_compare}
\caption{Cartpole training curve} \label{fig:compare-a}
\end{subfigure}\hfill
\begin{subfigure}{0.49\textwidth}
\centering     
\begin{tikzpicture}

\definecolor{crimson2411462}{RGB}{241,14,62}
\definecolor{darkgray176}{RGB}{176,176,176}
\definecolor{green}{RGB}{0,128,0}

\begin{axis}[
height=6cm,
tick align=outside,
tick pos=left,
width=7cm,
x grid style={darkgray176},
xlabel={Average absolute Q-value},
xmin=35.6929858737233, xmax=103.062238767918,
xtick style={color=black},
y grid style={darkgray176},
ylabel={Average reward},
ymin=92.3162058371736, ymax=245.037096774194,
ytick style={color=black}
]
\addplot [draw=blue, fill=blue, mark=*, only marks, opacity=0.5]
table{%
x  y
39.3025619601669 148.151851851852
39.4359016075954 149.816479400749
39.3473784246292 150.947169811321
39.9125303004682 136.057823129252
39.0490300317926 149.816479400749
40.7404618893005 116.961988304094
38.7552246416412 148.702602230483
39.2511665008046 129.873376623377
39.4601456129819 136.989726027397
39.266386630642 171.678111587983
39.4630991650723 133.782608695652
38.9621279328676 143.372759856631
39.3502184976771 152.095057034221
39.3462433474619 141.847517730496
39.6334884643361 138.892361111111
39.7162368741661 137.460481099656
39.1606728553839 138.892361111111
39.3907367990568 143.888489208633
39.7093587026432 133.336666666667
39.0470187431261 123.842105263158
40.1924534981579 158.106719367589
39.3541659688804 136.989726027397
39.7099019993171 159.366533864542
39.7226993586391 161.947368421053
39.2151235291831 160.646586345382
39.0238311851513 173.164502164502
39.1496581784816 141.847517730496
39.6168218800731 146.52380952381
39.2806327085808 163.938524590164
39.2779515443765 170.944444444444
};
\addplot [draw=black, fill=black, mark=*, only marks, opacity=0.5]
table{%
x  y
51.8447982086455 150.37969924812
52.0791065372124 135.138513513514
52.113207612893 145.458181818182
54.1333718683578 99.258064516129
53.3576080669716 139.376306620209
52.1947536908954 134.683501683502
52.3079692293651 151.518939393939
52.9373104477778 155.042635658915
52.0817992715239 162.605691056911
52.2342164899632 133.782608695652
52.2946379200537 166.670833333333
52.2564731393628 144.93115942029
52.1997107432664 161.947368421053
52.5293874831215 135.138513513514
51.7943861291824 190.480952380952
51.6654264584932 169.495762711864
52.7640217154294 132.016501650165
53.125697607556 148.702602230483
52.3835986580502 149.816479400749
52.460532563898 132.453642384106
52.6193761600412 139.863636363636
52.4744847605295 158.734126984127
53.7765719955891 148.702602230483
52.4151032639258 152.095057034221
52.0365373900548 140.354385964912
53.2360363316819 140.354385964912
52.4999887265235 148.151851851852
54.5055583420977 158.734126984127
53.4968275308911 117.304985337243
51.8347669138506 150.947169811321
};
\addplot [draw=green, fill=green, mark=*, only marks, opacity=0.5]
table{%
x  y
51.8222852764048 137.460481099656
52.3343709732652 142.35231316726
51.8848212961584 166.670833333333
51.866654731632 181.822727272727
52.5813304558313 144.93115942029
51.5005553994127 156.25390625
51.8522269068569 138.411764705882
52.0954045008764 138.892361111111
52.141458123868 125.394984326019
51.5455988932479 147.605166051661
51.8014834765423 141.847517730496
52.2701725127384 156.25390625
53.1497276931643 144.407942238267
53.0777455998205 143.372759856631
54.0964793840215 132.893687707641
52.4185798088543 131.150819672131
51.9604932565398 143.888489208633
52.4145760311822 160.004
51.9282453859001 144.93115942029
51.9200189678304 144.93115942029
51.2057654528653 158.106719367589
51.7473575813524 141.34628975265
51.9805878603879 125.789308176101
51.6985421384972 148.151851851852
51.7105377836119 162.605691056911
51.7223990182817 122.70245398773
51.961695134148 152.675572519084
51.8942537522808 161.29435483871
53.0265401395611 108.698369565217
51.6897877881851 152.675572519084
};
\addplot [draw=crimson2411462, fill=crimson2411462, mark=*, only marks, opacity=0.5]
table{%
x  y
53.1382784269443 163.934426229508
54.801753745042 144.927536231884
51.745071349404 152.091254752852
52.2429893382573 191.387559808612
52.4788034739107 182.648401826484
52.322190544048 185.185185185185
51.639544120086 173.913043478261
52.3506146260925 173.160173160173
53.5757508374564 160.642570281125
52.0159820921732 182.648401826484
51.9739450401165 212.765957446808
52.6028309701107 144.404332129964
52.012171808641 215.05376344086
52.3961330478277 183.48623853211
52.7361803044658 185.185185185185
52.4929489501115 176.211453744493
51.3248255539317 182.648401826484
52.1216617645323 166.666666666667
52.2354855247222 193.236714975845
52.3209972846709 184.331797235023
52.4837400361046 147.60147601476
52.0540810328871 238.095238095238
52.9053567497663 165.97510373444
52.5289769218884 176.211453744493
52.0977092141166 173.160173160173
51.906208710302 205.128205128205
52.9819895374238 160.642570281125
52.277717189493 176.991150442478
51.7467916965526 173.913043478261
52.299540485207 186.046511627907
};
\addplot [semithick, black, dashed]
table {%
100 92.3162058371736
100 245.037096774194
};
\end{axis}

\end{tikzpicture}
\caption{Cartpole reward vs. estimation analysis} \label{fig:compare-b}
\end{subfigure}

\begin{subfigure}{0.49\textwidth}
\centering     
\input{Acrobot_compare}
\caption{Acrobot training curve} \label{fig:compare-c}
\end{subfigure}\hfill
\begin{subfigure}{0.49\textwidth}
\centering     
\begin{tikzpicture}

\definecolor{crimson2411462}{RGB}{241,14,62}
\definecolor{darkgray176}{RGB}{176,176,176}
\definecolor{green}{RGB}{0,128,0}

\begin{axis}[
height=6cm,
tick align=outside,
tick pos=left,
width=7cm,
x grid style={darkgray176},
xlabel={Average $|Q|$-value},
xmin=0, xmax=100,
xtick style={color=black},
y grid style={darkgray176},
ylabel={Average reward},
ymin=-450.38885203175, ymax=-91.7471508115152,
ytick style={color=black}
]
\addplot [draw=blue, fill=blue, mark=*, only marks, opacity=0.5]
table{%
x  y
77.1669360449031 -126.458598726115
77.3330298487097 -131.960132890365
77.0021680849522 -140.921985815603
74.951901645489 -126.050793650794
73.0800520946011 -124.457680250784
70.7532851288199 -115.002898550725
79.3382559791148 -158.450199203187
74.8956938023016 -138.937062937063
71.5720622295648 -122.907120743034
77.856650811138 -127.684887459807
71.9951956433907 -121.021341463415
78.3983155242771 -139.922535211268
79.9325582657918 -150.598484848485
67.2417463588998 -113.345714285714
78.139095447436 -139.926056338028
71.1594563214436 -118.468656716418
80.1266632470399 -146.682656826568
76.1446529906049 -137.965277777778
83.1615791498587 -220.116022099448
80.5528829428986 -168.584745762712
76.7815032940865 -139.431578947368
73.7424156853095 -125.651898734177
70.8466138398625 -119.548192771084
81.1307845104337 -163.699588477366
74.269726441358 -126.053968253968
79.9033560012728 -148.895131086142
68.3648582737282 -116.020467836257
79.4738803186707 -157.81746031746
67.3042050512761 -112.693181818182
81.2453609858751 -164.380165289256
};
\addplot [draw=black, fill=black, mark=*, only marks, opacity=0.5]
table{%
x  y
65.7107016790092 -130.648026315789
68.0037603492558 -130.219672131148
68.2440700667441 -159.088
69.9076334176168 -175.303964757709
67.5473768457875 -130.648026315789
68.081295200789 -144.007246376812
67.4157192988098 -132.852842809365
68.1185433421984 -159.730923694779
65.0583470086798 -123.291925465839
64.1444124869891 -126.458598726115
67.0094834406406 -132.406666666667
68.0218284050718 -131.082508250825
67.3944819811627 -138.933566433566
68.2608020928979 -182.596330275229
65.6805821869701 -126.053968253968
68.5302748423256 -192.347826086957
69.3075773464039 -208.539267015707
67.6009304096505 -157.81746031746
66.7050589160919 -127.688102893891
67.085341365321 -167.869198312236
64.475303810589 -116.023391812865
68.7815392960534 -160.379032258065
66.8009513627961 -139.926056338028
69.0301169711694 -178.470852017937
69.7406805001445 -163.02868852459
65.9507326008365 -152.930769230769
68.2914370252565 -167.163865546218
67.1398611397028 -128.521035598706
68.8910564876229 -173.008695652174
66.1106191979662 -131.963455149502
};
\addplot [draw=green, fill=green, mark=*, only marks, opacity=0.5]
table{%
x  y
71.4387179823294 -171.504310344828
63.259672358422 -123.679127725857
68.4540243353725 -145.06204379562
66.4235995737821 -135.59385665529
65.9088424261421 -144.007246376812
64.2669349113777 -125.648734177215
68.4138836250186 -140.427561837456
68.8408077901855 -157.813492063492
68.3840900235623 -140.921985815603
67.4964090169072 -131.085808580858
68.7251184222654 -153.517374517375
66.2135595998347 -127.275641025641
72.6229113805957 -172.255411255411
66.3414560289562 -148.898876404494
60.9810573473662 -124.460815047022
69.8999146573752 -144.534545454545
64.0778988284662 -125.651898734177
68.5828488186717 -137.484429065744
62.9454766708195 -125.648734177215
68.3414192892045 -149.458646616541
71.800783943674 -160.379032258065
72.8339757610053 -215.335135135135
67.9804153438002 -130.651315789474
66.3536572180241 -136.53264604811
64.935029422453 -127.688102893891
69.3515581968576 -146.139705882353
67.127249760361 -128.521035598706
68.4485784221023 -135.12925170068
66.7034340974629 -140.918439716312
72.8452020678416 -192.342995169082
};
\addplot [draw=crimson2411462, fill=crimson2411462, mark=*, only marks, opacity=0.5]
table{%
x  y
85.1360123744607 -151.751908396947
82.7828237935424 -127.684887459807
72.9742866926759 -114.664739884393
69.2152074534535 -108.049046321526
71.7642063218743 -117.762611275964
77.0063308478907 -117.05604719764
75.707817283231 -127.688102893891
90.1049461709581 -234.417647058824
79.8253484727412 -128.944805194805
88.8462948001295 -249.15
72.9956557542115 -130.648026315789
68.3662480779976 -109.555248618785
70.0287333279029 -109.861495844875
89.6844371589184 -260.607843137255
83.3945883918494 -136.058219178082
84.6927877131566 -143.494584837545
80.6348654646121 -159.730923694779
72.8721807329252 -116.360703812317
83.2234592260331 -141.935714285714
90.9847751316935 -434.086956521739
64.2679866075695 -109.861495844875
89.0938635350794 -356.392857142857
65.3365931466967 -112.050847457627
87.0988399514228 -244.564417177914
77.7897523649991 -119.909365558912
75.432720368509 -119.545180722892
78.9175407677323 -147.244444444444
68.2732234408811 -109.250688705234
87.3975154962659 -171.5
79.9345115423203 -150.598484848485
};
\addplot [semithick, black, dashed]
table {%
100 -450.38885203175
100 -91.7471508115152
};
\end{axis}

\end{tikzpicture}
\caption{Acrobot reward vs. estimation analysis} \label{fig:compare-d}
\end{subfigure}

\begin{subfigure}{0.49\textwidth}
\centering     
\input{mc_compare}
\caption{Mountain Car training curve} \label{fig:compare-e}
\end{subfigure}\hfill
\begin{subfigure}{0.49\textwidth}
\centering     
\begin{tikzpicture}

\definecolor{crimson2411462}{RGB}{241,14,62}
\definecolor{darkgray176}{RGB}{176,176,176}
\definecolor{green}{RGB}{0,128,0}

\begin{axis}[
height=6cm,
legend cell align={left},
legend style={fill opacity=1, draw opacity=1, text opacity=1, at={(0.97,0.03)}, anchor=south east},
log basis x={10},
tick align=outside,
tick pos=left,
width=7cm,
x grid style={darkgray176},
xlabel={Average $|Q|$-value (log)},
xmin=50.9604756063092, xmax=21809.2710083462,
xmode=log,
xtick style={color=black},
xtick={1,10,100,1000,10000,100000,1000000},
xticklabels={\ensuremath 1,10,100,1000,10000,100000,1000000},
y grid style={darkgray176},
ylabel={Average reward},
ymin=-1042.51847005988, ymax=-107.185462075848,
ytick style={color=black}
]
\addplot [draw=blue, fill=blue, mark=*, only marks, opacity=0.5]
table{%
x  y
91.3147050864845 -372.209677419355
6029.72955144539 -210.52701754386
1429.41154502367 -246.306239737274
92.5497258932378 -204.778839590444
1843.76216553083 -240.192954363491
716.420404058026 -237.530482977039
264.904897502357 -203.11509817197
291.277937068251 -296.443675889328
2626.44705633034 -222.883358098068
7237.22828125973 -213.675925925926
111.210162865501 -243.112641815235
136.287629781858 -219.138787436085
86.3127726650742 -292.683902439024
172.776733237679 -262.238636363636
94.379392732589 -852.275568181818
90.0365657105718 -590.553149606299
103.03518314739 -328.228665207877
107.255857875527 -319.830490405117
93.0521781732905 -427.961483594864
321.350489239983 -259.966204506066
3060.13217604341 -194.049805950841
1769.29679112701 -233.282270606532
3187.68118629312 -231.661003861004
94.3846925855236 -946.375394321767
75.6609349318418 -277.265249537893
767.473282987791 -297.62003968254
2830.18004317522 -214.286428571429
118.523754369156 -241.741337630943
126.850275219765 -255.102891156463
94.8636979112752 -279.590866728798
};
\addlegendentry{Avg DQN}
\addplot [draw=black, fill=black, mark=*, only marks, opacity=0.5]
table{%
x  y
109.962662358181 -215.363244795406
16558.9086139251 -185.874225526642
75.1264300180001 -201.342953020134
859.641626400496 -193.741430700447
520.733161609965 -255.537478705281
1806.16350289864 -169.683823529412
1136.5880064261 -175.644613583138
587.456572343173 -217.392028985507
5295.38594887539 -173.712217718587
83.2526190541965 -236.780584056827
128.040569121769 -324.325405405405
7301.39632183378 -184.049693251534
227.784219126798 -242.915789473684
906.313691052888 -251.678691275168
95.9241994879023 -872.095930232558
538.868994449864 -258.176419965577
1431.83721247973 -186.567786069652
2134.20332315799 -251.890008396306
77.0058686223579 -242.71925566343
4546.40588786596 -184.616
84.6664374150422 -204.639154160982
123.104270154029 -330.034103410341
559.283498898714 -209.35170969993
4961.33984211051 -172.513513513514
431.577274285755 -207.326192121631
1448.46761177852 -224.383694839192
77.4362606189187 -208.768963117606
634.137507439096 -243.507305194805
90.7181992409742 -385.110397946085
93.7820611754572 -465.840062111801
};
\addlegendentry{DQN}
\addplot [draw=green, fill=green, mark=*, only marks, opacity=0.5]
table{%
x  y
99.1835888900338 -1000.00333333333
93.3294798455113 -387.098064516129
507.405163874699 -203.11509817197
2095.00640414419 -191.939219449776
300.88651944473 -319.15
5518.97642992821 -173.110790536642
99.4650963422679 -265.252873563218
95.1088169036155 -308.009240246407
10918.2819913696 -189.155737704918
359.768568568661 -197.369078947368
122.747533936843 -209.937718684395
4007.92372235173 -192.555198973042
1307.20965707946 -216.294881038212
2057.52859535057 -215.672897196262
3144.83411130501 -172.414367816092
89.680130601973 -244.101708706265
667.478213094388 -208.9143454039
558.110044515287 -204.08231292517
2256.40672523169 -216.607220216606
103.002351523039 -585.939453125
5432.14846808023 -185.759133126935
9478.02659168002 -205.480136986301
983.237005448485 -256.850171232877
99.1772188926404 -974.029220779221
921.490080801333 -207.183011049724
99.1954789526934 -946.375394321767
94.4592547367913 -253.807952622673
4251.22912789453 -233.645638629283
335.713973888153 -251.467728415759
3459.92046098545 -188.088401253919
};
\addlegendentry{Double DQN}
\addplot [draw=crimson2411462, fill=crimson2411462, mark=*, only marks, opacity=0.5]
table{%
x  y
87.4032555854313 -325.732899022801
75.1948773181905 -187.5
70.7116194960881 -156.168662155128
93.4171620706616 -607.287449392713
93.2829274051342 -480.769230769231
78.4516761894558 -207.03933747412
75.6048347126083 -180.722891566265
93.6626190212353 -993.377483443709
80.0346159353772 -232.018561484919
67.1186035942963 -149.700598802395
76.7258400738423 -185.873605947955
74.5741074135739 -172.910662824207
72.8757526499278 -163.934426229508
82.8780591135395 -232.018561484919
76.3897203204122 -193.673337637185
76.3611757391264 -166.389351081531
81.7648031487853 -209.643605870021
93.4038225483601 -940.438871473354
102.536748023094 -225.225225225225
88.5638910839599 -271.24773960217
78.1925209714585 -192.926045016077
70.0702995542806 -157.068062827225
72.3311814797188 -165.562913907285
93.3047903841628 -625
74.3303458545993 -186.219739292365
81.9122678879437 -207.325501036628
80.9813911031501 -202.292650033715
81.1948163409422 -262.46719160105
91.6590520335263 -375
93.350749736845 -472.44094488189
};
\addlegendentry{ES DQN}
\addplot [semithick, black, dashed, forget plot]
table {%
100 -1042.51847005988
100 -107.185462075848
};
\end{axis}

\end{tikzpicture}
\caption{Mountain Car reward vs. estimation analysis} \label{fig:compare-f}
\end{subfigure}

\caption{Here the Elastic Step DQN algorithm is compared across the three test bed environments against well known variants of DQN that are known to address the divergent and overestimation  issue. (a), (c) and (e) plots the training curve for each of the tested environments while (b), (d) and (f) plots the relationship between reward against the $|Q|$-value estimate.  The dotted dash vertical line represents the upper bound of the absolute $|Q|$-value.} \label{fig:compare}
\end{figure*}

\begin{table}
\begin{center}
\caption{ This table outlines the final training reward performance of each algorithm averaged over the 30 seeds. The final reward performance is defined as only the last 10 epochs of each run. A two sample t-test is then calculated on the final performance data between Elastic Step DQN and all benchmark algorithms with the p-value reported for each environment in the table. The algorithms with the highest reward are in bold text}
\begin{tabular}{llllllllll}

& \multicolumn{3}{l}{Cartpole} & \multicolumn{3}{l}{Acrobot} & \multicolumn{3}{l}{Mountain Car} \\ 
\hline
& {$\bar{x}$ reward} & $s$ & $p$-value   & { $\bar{x}$ reward} & $s$    & $p$-value   & { $\bar{x}$ reward} & $s$ & $p$-value   \\
\hline
\\
DQN &	100.7 &	29.6 &	$<$0.05 &	-107.6 & 89.4 &	0.22 &	-156.8 &	26.5 &	$<$0.05 \\
2 step &	74.3 &	31.3 &	$<$0.05 & -121 &	89.4 &	0.12 & -462.9 &	153.5 &	$<$0.05 \\
3 step &	76.9 &	40.7 &	$<$0.05 & -110.6 & 28.9 &	$<$0.05 &	-880.8 &	161.5 &	$<$0.05 \\
4 step &	29 &	20.5 &	$<$0.05 & -93.2 &	20.6 &	0.75 &	-753.8 &	180 &	$<$0.05 \\
6 step &	9.8 &	1.2 & $<$0.05 & \textbf{-91.5} & 16.4 &	0.39 &	-513.4 & 226.5 & $<$0.05 \\
8 step &	10 &	1.5 &	$<$0.05 & -213.8 & 70.4 &	$<$0.05  &	-891.4 & 191 &	$<$0.05 \\
Double &	225.7 &	166.6 &	$<$0.05 &	-103.3 & 19 &	0.19  &	-162.8 & 55.6 &	$<$0.05 \\
Average &	183.6 &	38.2 &	$<$0.05 &	-100.8 & 31.4 &	0.31  &	-153.9 & 21.7 &	$<$0.05 \\ 
Elastic Step &	\textbf{413.4} &	112.2 &	&	-94.5 &	9.9 & & \textbf{-123.4} &	19.8 &	\\

\hline

\end{tabular}

\label{tab:stat-sig}\end{center}
\end{table}

\subsection{Analysis of the final training performance across $n$-step variants and DQN extensions}

Elastic Step DQN was developed on the principle that dynamic selection of the value $n$ can reduce overestimation  bias and improve algorithmic training performance. The empirical trials computed across Cartpole, Acrobot and Mountain Car consistently showed statistically significant higher final training performance against a majority of the $n$-step update algorithms. Figure \ref{fig:nstep-a}, \ref{fig:nstep-c} and \ref{fig:nstep-e} compares the training results between Elastic DQN against the $n$-step variants $(2/3/4/6/8)$ measured in terms of the average rewards ($\bar{x}$) received per epoch along with the standard deviation ($s$) of each epoch also plotted on the graph. Cartpole has been acknowledged \citep{ceron2021revisiting} to be very sensitive to hyper-parameter tuning and in our experiments shown itself to be prone to divergent behaviour -- as a result it is a good environment to test for training efficacy. In the Cartpole experiments, Elastic Step DQN took an average of 1.05 steps (with a maximum of 297 and a minimum of 1 step per update), and was able to achieve statistically significant higher average final training performance ($\bar x$ reward = 413.4) than the $n$-step variants as well as single-step DQN ($\bar x$ reward = 100.7. Observations from Figure \ref{fig:nstep-a}, showed that while $n$-step variants exhibited stronger early stage performance in the environments, a majority of the $n$-step variants struggled to learn any meaningful policy in the Cartpole environment. The optimal value of $n$ for the Cartpole environment was 3, which peaked after the 33rd epoch and suffered periodic degradation of training performance over time. This periodic degradation of training performance can also be observed in single step DQN which went through a period of stability between the 25th epoch to the 70th epoch before collapsing afterwards. This instability is again observed in Double DQN in Figure \ref{fig:compare-a} but not Average DQN. Interestingly, when the termination condition was set much higher in the Cartpole environment, both Average DQN and Double DQN rarely achieved rewards above 200 which was the original default termination condition, while Elastic Step DQN was able to consistently surpass the 200 rewards mark. In the experiments against the DQN extenstions, Elastic Step DQN ($\bar x$ reward = 413.4), was able to achieve statistically significant improvements against Double DQN ($\bar x$ reward = 225.7) and Average DQN ($\bar x$ reward = 183.6) as indicated in Table \ref{tab:stat-sig}. 

Of the three environments, Acrobot was the easiest environment for all the algorithmic implementations. The $n$-step variants performed comparatively well in the Acrobot environment, both 4/6-both (4-step: $\bar x$ reward = -93.2, 6-step: $\bar x$ reward = -91.5) achieved higher average final training performance than Elastic Step DQN ($\bar x$ reward = -94.5). In spite of this, the training curves in Figure \ref{fig:nstep-c} does show that Elastic Step DQN converged much faster than the $n$-step variants and for a majority of the epochs had sustained higher rewards than the $n$-step. In this environment, statistically significant improvements was achieved against 3 and 8-step DQN. Single-step DQN ($\bar x$ reward = -107.6), Average ($\bar x$ reward = -100.8) and Double DQN ($\bar x$ reward = -103.3) also had very little issue converging to a decent policy and statistical significance could not be claimed against either. In this environment Elastic Step DQN took an average of 8.59 steps during its run (with a maximum of 923 steps and a minimum of 1 step per updates), demonstrating that this algorithm does indeed adjust the step-size to suit the environment.

The Mountain Car environment proved also to be a difficult for $n$-step variants which generally have learned sub-optimal policies. Figure \ref{fig:nstep-c} plots the training curve for the $n$-step variants, single step DQN and Elastic Step DQN. The plot shows that the $n$-step variants suffered from highly volatile learning with most collapsing after a period of learning like the training curves observed in Figure \ref{fig:nstep-a}. From Figure \ref{fig:nstep-a} it seems lower step sizes introduce more stability $2/4$-step DQN suffer from a much less dramatic collapse of the learning than $6/8$-step. On the contrary, Elastic Step DQN took an average of 7.22 steps (with a maximum of 1000 steps and a minimum of 1 step per update) to converge to an optimal policy. In these experiments, Elastic Step DQN achieved statistically significant improvements ($\bar x$ reward = -123.4) against the $n$-step variants (2-step DQN $\bar x$ reward = -462.9) and single step DQN ($\bar x$ reward = -156.8). $n$-step variants suffered from highly learning volatility as observed from the standard deviation of the rewards which were at least three times the standard deviation of single-step DQN and five times that of Elastic Step DQN (refer to Table \ref{tab:stat-sig}). Close inspection into single seed runs of the Mountain Car environment have shown instances where both the 2 and 6 step DQN implementations have outperformed the Elastic Step DQN and single-step DQN algorithm by moderate margins. These results run in contrary to the portrayed image that multi-step DQN are in general more performant than single-step DQN. It is possible that perhaps the step size selection was not suitable, however this validates how difficult it may be to select the correct horizon. When compared against the DQN extensions, we saw that Average DQN and Double DQN algorithm have been able to converge to an optimal policy much faster than Elastic Step DQN in Figure \ref{fig:compare-c}, but on average achieved a statistically significant lower final training performance (Table \ref{tab:stat-sig}: Double DQN $\bar x$ reward = -162.8, Average DQN $\bar x$ reward = -153.9) than Elastic Step DQN ($\bar x$ reward = -123.4). 

One noticeable trend in Figures \ref{fig:compare-a}, \ref{fig:compare-c} and \ref{fig:compare-e} is that Elastic Step DQN is slower to converge to an optimal policy than Double DQN and Average DQN. This slower convergence speed could be attributed to the large step sizes the agent needs to take in the early stages of learning when the data is scarce and the clustering algorithm does not have representative data sample to train on. As the agent accumulates more data, the agent naturally takes smaller step sizes. In spite of this disadvantage, Elastic Step DQN has shown comparable performance against the $n$-step variants and well established DQN extensions. Whether this is a result of more accurate $Q$ value estimations will be explored in Section \ref{sec:diverge}.
    
\begin{table}[]
\begin{center}

\caption{Summary statistics steps taken during Elastic Step DQN runs.}

\begin{tabular}{llll}
\hline
& Cartpole & Acrobot & Mountain Car \\ \hline
{ $\bar{x}$ steps}    & 1.05     & 8.59    & 7.22         \\
{$Min$ steps}     &  1     & 1       & 1            \\
{$Max$ steps}     & 297        & 923     & 1000         \\
{$x_{0.5}$ steps}  & 1        & 3       & 2            \\
{$s$ steps}    & 0.84     & 18.99   &  21.11  \\
\hline
\end{tabular}
\label{summ-runs}\end{center}
\end{table}
\subsection{Analysis of overestimation  and soft divergence} \label{sec:diverge}

\begin{table}
\begin{center}
\caption{This table outlines the average absolute $Q$ value estimate ($\bar{x}$ $|Q|$), the standard deviation of the absolute $Q$ value estimate ($s$ $|Q|$) and the median absolute $Q$ value estimate ($x_{0.5}$) of each algorithm over the 30 seeds. $|Q|$-values that exceed 100 are in bold.}
\begin{tabular}{llllllllll}
\hline
 & \multicolumn{3}{l}{Cartpole} & \multicolumn{3}{l}{Acrobot} & \multicolumn{3}{l}{Mountain Car}  \\
 & $\bar{x}$  $|Q|$    & $s$    & $x_{0.5}$    & $\bar{x}$  $|Q|$    & $s$     & $x_{0.5}$    & $\bar{x}$  $|Q|$    & $s$     & $x_{0.5}$ \\
\hline
DQN &	52.6 &	24.1 &	52.4 &	67.4 &	22.8 &	67.6 &	\textbf{1803.1} &	3445.6 &	\textbf{549.1} \\
2 step DQN &	33.3 &	17.9 &	33.1 &	65 &	19.9 &	66.5 &	98.1 &	10.4 &	98.4 \\
3 step DQN &	43.7 &	22.2 &	42.6 &	39.5 &	19.6 &	39.4 &	98.6 &	10.1 &	98.6 \\
4 step DQN &	56 &	32.4 &	43.9 &	79.9 &	22.2 &	80.3 &	98.3 &	10.7 &	98.4 \\
6 step DQN &	\textbf{415.2} &	722.5 &	56.7 &	92.1 &	13.6 &	92.7 &	98.5 &	10.6 &	98.5 \\
8 step DQN &	\textbf{2919.6} &	5634.7 &	\textbf{278.4} &	97.3 &	4.4 &	97.6 &	98.7 &	10.4 &	98.7 \\
Double DQN &	52.1 &	24 &	51.9 &	67.7 &	21.3 &	68.2 &	\textbf{1985.2} &	2701.8 &	\textbf{612.8} \\
Average DQN &	39.4 &	19.4 &	39.3 &	75.8 &	14.7 &	76.9 &	\textbf{1135.5} &	2065 &	\textbf{154.5} \\
Elastic Step DQN &	52.4 &	23.2 &	65.6 &	78.8 &	13.1 &	79.4 &	81.9 &	14.9 &	79.4 \\

\hline
\end{tabular}
\label{tab:q-data} \end{center}
\end{table}

\begin{table}
\caption{Maximum average absolute $Q$ value estimate (max $|Q|$), and the Spearman correlation between training reward and $|Q|$ of each algorithm over the 30 seeds. $|Q|$-values that exceed 100 are in bold.}
\begin{tabular}{lllllll}
\hline
& \multicolumn{2}{l}{Cartpole}   & \multicolumn{2}{l}{Acrobot}   & \multicolumn{2}{l}{Mountain Car} \\  
& Spearman  & max $|Q|$ & Spearman & max $|Q|$ & Spearman  & max $|Q|$ \\
& correlation  &   & correlation &  & correlation  &   \\
\hline
DQN &	-0.4 &	54.5 &	-0.8 &	69.9 &	0.5 &	\textbf{16558.9} \\
2 step &	0 &	44.1 &	-0.5 &	71.2 &	-0.1 &	98.8 \\
3 step &	0 &	66.8 &	0.2 &	40.3 &	-0.4 &	99.2 \\
4 step &	-0.3 &	\textbf{153.9} &	-0.1 &	84.4 &	0.1 &	98.9 \\
6 step &	-0.6 &	\textbf{5455.3} &	-0.5 &	94 &	-0.1 &	98.9 \\
8 step &	-0.8 &	\textbf{31125.4} &	-0.3 &	98.1 &	0.5 &	99.1 \\
Double &	54.1 &	-0.4 &	-0.9 &	72.8 &	0.8 &	\textbf{10918.3} \\
Average &	-0.2 &	40.7 &	-1 &	83.2 &	0.7 &	\textbf{7237.2} \\
Elastic Step &	-0.4 &	81.4 &	0.9 &	91 &	-0.9 &	\textbf{102.5} \\

\hline
\end{tabular}
\label{tab:corr}
\end{table}

Given Elastic Step DQN's strong final training performance observed in Section \ref{sec:extset}, the question is then, is this as a result of the ability of Elastic Step DQN to moderate $Q$ value estimates? As discussed in Section \ref{sec:extset}, we have established that $|Q|$-value estimates greater than 100 can be considered to be suffering from overestimation. Based on this definition, based on the mean $|Q|$-value estimates from Table \ref{tab:q-data}, Elastic Step DQN did not suffer from overestimation, although examination of the max $|Q|$ from the Mountain Car environment in Table \ref{tab:corr} indicates that it breached the $|Q| > 100$ threshold by 2.5. However, this value is negligible when compared to single-step DQN, Double DQN and Average DQN which had an average $|Q|$-value estimate of 1803.1, 1,985.2 and 1,135.5 respectively in the Mountain Car environment.  Interestingly, despite the overestimation, Figure \ref{fig:compare-f} seem to show that the runs where the $|Q|$-value estimates were above 100 did very well while runs that were within the [0, 100] bound had mix performances. When we observe Figure \ref{fig:nstep-b}, \ref{fig:nstep-d}, and \ref{fig:nstep-f}  for the three environments across the $n$-step DQN, we can observe that there exist a visual relationship between poor performance and a overestimated $|Q|$ value. Elastic Step DQN tended to cluster together where the $|Q|$ values are lower and the reward is higher, on the other hand, the $n$-step algorithms had very volatile $|Q|$ values, and the higher $|Q|$ values also visually aligned with poorer performance. In contrast, we saw a contradictory relationship in Figure  \ref{fig:compare-f}, where despite DQN, Double DQN and Average DQN crossed the upper bound limit, it consistently performed well. The Spearman correlation as stated in Table \ref{tab:corr} for single-step DQN, Double DQN and Average DQN were 0.5, 0.8 and 0.7 respectively which indicated there existed a moderate to strong monotonic relationship between reward and $|Q|$-value. An increase in the $|Q|$-value estimates above the upper bound of 100 seem to positively correlate which stronger training performance which seems to indicate that overestimation  does not actually guarantee poor performance or instability. One explanation could be that overestimation  is only bad when the relative value of the state action value estimates do not encourage optimal action. In \citet{lan2019maxmin} analysis of overestimation indicated that overestimation can be harmful or beneficial depending on the environment. Of course, there were situations where overestimations did correlate with poor performance -- in the case of Cartpole, each increase of the step size in the value of $n$ lead to incrementally higher $|Q|$-value estimates -- both 6, and 8-step DQN experienced tendencies to over-estimate. Overestimation in multi-step DQN have only been recorded when the step-size horizon was incorrectly tuned, which seems to confirm the idea that correctly tuned multi-step algorithms are able to moderate overestimation  bias. In some cases the $|Q|$-value estimates were relatively lower than well-performing algorithms which could be potentially also be harmful as underestimation as explored by \citep{he2020wd3, lan2019maxmin} indicated that agents who were prone to underestimation bias tended to incur non-optimal performance.

\subsection{Cluster analysis of the Mountain Car environment}

\begin{figure}
\centering
  \centering
  \includegraphics[width=12cm]{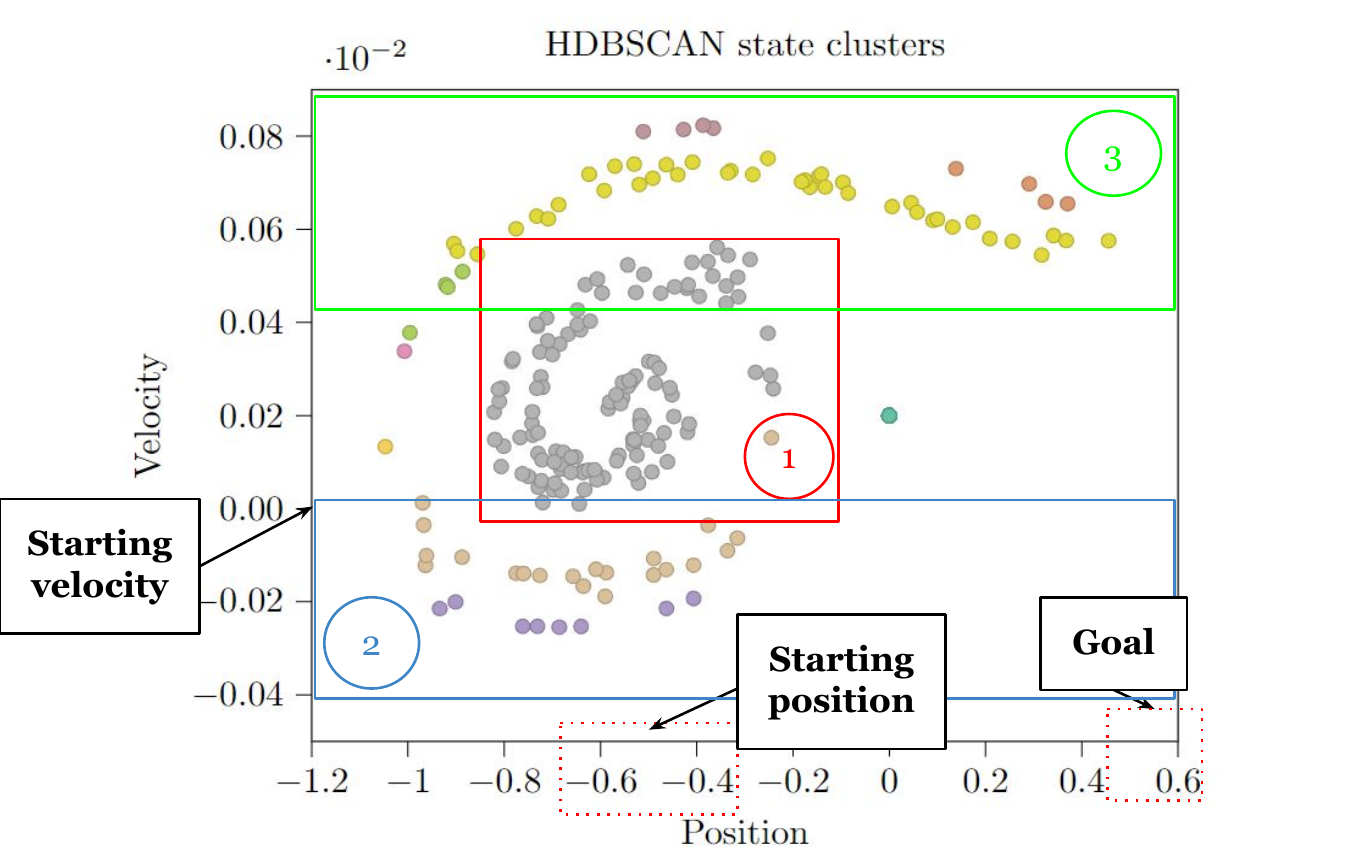}
  \label{clusters}
\caption{The scatter plot contains observations extracted from HDBSCAN during a single Mountain Car run. The car can start at any point between the x-axis -0.6 and -0.4, while position 0.5 represents the goal region. Each colour represents a seperate cluster identified by HDBSCAN.}
\label{fig:clustering-fig}
\end{figure}

The primary objective of the unsupervised clustering algorithm was to leverage its mechanism of similarity detection to assist with the selection of the step size. To validate whether the clusters are rationally assigned, there needs to be a close inspection into the clusters generated from a successful run of the algorithm. The Mountain Car environment is a good environment to analyse the clustering performance because the OpenAI implementation exposes a two element vector state space which can easily be mapped onto a two dimensional plot. Figure \ref{fig:clustering-fig} displays the cluster results mapped against the observational vectors from a single successful run of Mountain Car, each colour represents a different set of cluster labels and each data point represents a single observation vector. The x-axis is represented by the position of the car at a certain point in time while the y-axis is represented by velocity that the car was travelling at (a negative value indicates the car is moving backwards while a positive value indicates the car is moving towards the goal). The results from the clusters shows that the unsupervised clustering algorithm systematically identified seven main clusters which are all uneven and dense separated by areas of low density data points. Observations into the individual clusters indicate that HDBSCAN has identified and separately clustered varying state action strategies together. To beat the Mountain Car environment, the agent needs to move back and forth at the bottom of the mountain to gain momentum this is labelled 1 in figure \ref{fig:clustering-fig}. As the car reached middle of the mountain, the momentum builds up and the car gains more velocity over time as indicated by label 2 and 3. These clusters identified by HDBSCAN are inline with the expected behaviour of the algorithm.

\section{Conclusion and future work}

Overestimation  in the $|Q|$-value estimates has been an enduring challenge in Deep Q-networks and overestimation s can stifle learning and lead the agent to learn sub-optimal polices. This paper takes a fresh look at two ideas prevalent in the literature that may potentially address the overestimation  issue. The first idea centres around a school of thought that multi-step updates can alleviate the overestimation  pressures in deep value based reinforcement learning--this idea has been extensively experimented with in the literature and have shown promise. However the application of the approach can be difficult because multi-step updates can be very sensitive to the hyper-parameters. The second idea centres around Coarse Q-learning and the idea that systematic generalisation of similar states can alleviate divergent behaviour. 

In this paper a new algorithm called Elastic Step DQN is presented that combines the ideas mentioned to address overestimation  in Deep Q-Networks. The Elastic Step Agent utilises a clustering algorithm to systematically discern states that are similar together and belong to a single multi-step update. This allows the agent to automate the step size and leverage the benefits of multi-step updates. The empirical analyses in the paper so far has shown that Elastic Step DQN can on average achieve comparable or better results than $n$-step variants and established DQN extensions while moderate overestimation  in $|Q|$-value estimates. 

One of the disadvantages of Elastic Step DQN is that it is slow in the early stages of learning to converge to an optimal policy, in spite of this it has shown to achieve comparable or better final performance to Average DQN and Double DQN. In addition, observed behaviours of overestimation  in Double DQN and Average DQN did not guarantee poor performance and in fact experience consistently stronger performance than instances where the $|Q|$-value was estimated within the bounds. In the same vein, a desirable estimation of the Q value does not mean that the algorithm will perform well.

These results highlights its potential for further exploration, So far the environments used in this paper have been small control problems, with vectored inputs, it would be interesting for future work to expand the experimental analyses into larger pixel based environments and perhaps apply the ideas to actor critic based algorithms and explore its impacts. It would also be beneficial to understand whether the algorithm could add incremental benefits when combined with other approcahes. Additionally, there is also potential to explore new methods to discern similarity that is more computationally efficient and more accurate.

\bibliographystyle{cas-model2-names}

\bibliography{cas-refs}

\appendix
\section{Appendix A}

\subsection{Hyper-parameters}
\onecolumn
\begin{longtable}[c]{@{}lllll@{}}
\multicolumn{1}{c}{Algorithm}                    & Hyper-parameter       & \multicolumn{1}{c}{Cartpole} & \multicolumn{1}{c}{Acrobot} & Mountain Car \\* 

\hline

\multicolumn{1}{l}{\multirow{12}{*}{Average DQN}} & learning  rate          & 0.0001 &  0.003  & 0.002\\
\multicolumn{1}{c}{}         & target update horizon   & 250       & 20        & 250          \\
\multicolumn{1}{c}{}         & replay capacity         & 10000     & 10000     & 10000     \\
\multicolumn{1}{c}{}         & initial replay size     & 500       & 500       & 500       \\
\multicolumn{1}{c}{}         & train frequency         & 1         & 1         & 1         \\
\multicolumn{1}{c}{}         & gamma                   & 0.99      & 0.99      & 0.99      \\
\multicolumn{1}{c}{}         & epsilon min             & 0.1       & 0.1       & 0.1       \\
\multicolumn{1}{c}{}         & epsilon start           & 1         & 1         & 1         \\
\multicolumn{1}{c}{}         & epsilon decay strategy  & linear    & linear    & linear    \\
\multicolumn{1}{c}{}         & batch size              & 32        & 32        & 32        \\
\multicolumn{1}{c}{}         & hidden units            & 200       & 24        & 24        \\
\multicolumn{1}{c}{}         & number of approximators & 2         & 2         & 2         \\
\hline
\multirow{11}{*}{Double DQN} & learning  rate          & 0.0001    & 0.003     & 0.0005    \\
                             & target update horizon   & 250       & 100       & 20        \\
                             & replay capacity         & 10000     & 10000     & 10000     \\
                             & initial replay size     & 500       & 500       & 500       \\
                             & train frequency         & 1         & 1         & 1         \\
                             & gamma                   & 0.99      & 0.99      & 0.99      \\
                             & epsilon min             & 0.1       & 0.1       & 0.1       \\
                             & epsilon start           & 1         & 1         & 1         \\
                             & epsilon decay strategy  & linear    & linear    & linear    \\
                             & batch size              & 32        & 32        & 32        \\
                             & hidden units            & 200       & 24        & 24        \\
\hline
\multirow{11}{*}{2-step DQN} & learning  rate          & 0.00025   & 0.0001    & 0.0001    \\
                             & target update horizon   & 1000      & 100       & 20        \\
                             & replay capacity         & 10000     & 10000     & 10000     \\
                             & initial replay size     & 500       & 500       & 500       \\
                             & train frequency         & 1         & 1         & 1         \\
                             & gamma                   & 0.99      & 0.99      & 0.99      \\
                             & epsilon min             & 0.1       & 0.1       & 0.1       \\
                             & epsilon start           & 1         & 1         & 1         \\
                             & epsilon decay strategy  & linear    & linear    & linear    \\
                             & batch size              & 32        & 32        & 32        \\
                             & hidden units            & 512       & 24        & 24        \\
\hline
\multirow{11}{*}{4-step DQN} & learning  rate          & 0.00025   & 0.0001    & 0.0001    \\
                             & target update horizon   & 1000      & 100       & 100       \\
                             & replay capacity         & 10000     & 10000     & 10000     \\
                             & initial replay size     & 500       & 500       & 500       \\
                             & train frequency         & 1         & 1         & 1         \\
                             & gamma                   & 0.99      & 0.99      & 0.99      \\
                             & epsilon min             & 0.1       & 0.1       & 0.1       \\
                             & epsilon start           & 1         & 1         & 1         \\
                             & epsilon decay strategy  & linear    & linear    & linear    \\
                             & batch size              & 32        & 32        & 32        \\
                             & hidden units            & 512       & 24        & 24        \\
\hline
\multirow{11}{*}{6-step DQN} & learning  rate          & 0.00025   & 0.001     & 0.0001    \\
                             & target update horizon   & 1000      & 100       & 100       \\
                             & replay capacity         & 10000     & 10000     & 10000     \\
                             & initial replay size     & 500       & 500       & 500       \\
                             & train frequency         & 1         & 1         & 1         \\
                             & gamma                   & 0.99      & 0.99      & 0.99      \\
                             & epsilon min             & 0.1       & 0.1       & 0.1       \\
                             & epsilon start           & 1         & 1         & 1         \\
                             & epsilon decay strategy  & linear    & linear    & linear    \\
                             & batch size              & 32        & 32        & 32        \\
                             & hidden units            & 512       & 24        & 24        \\
\hline
\multirow{11}{*}{8-step DQN} & learning  rate          & 0.00025   & 0.003     & 0.0001    \\
                             & target update horizon   & 1000      & 20        & 100       \\
                             & replay capacity         & 10000     & 10000     & 10000     \\
                             & initial replay size     & 500       & 500       & 500       \\
                             & train frequency         & 1         & 1         & 1         \\
                             & gamma                   & 0.99      & 0.99      & 0.99      \\
                             & epsilon min             & 0.1       & 0.1       & 0.1       \\
                             & epsilon start           & 1         & 1         & 1         \\
                             & epsilon decay strategy  & linear    & linear    & linear    \\
                             & batch size              & 32        & 32        & 32        \\
                             & hidden units            & 512       & 24        & 24        \\
\hline
\multirow{11}{*}{DQN}        & learning  rate          & 0.0001    & 0.001     & 0.0005    \\
                             & target update horizon   & 250       & 100       & 100       \\
                             & replay capacity         & 10000     & 10000     & 10000     \\
                             & initial replay size     & 500       & 500       & 500       \\
                             & train frequency         & 1         & 1         & 1         \\
                             & gamma                   & 0.99      & 0.99      & 0.99      \\
                             & epsilon min             & 0.1       & 0.1       & 0.1       \\
                             & epsilon start           & 1         & 1         & 1         \\
                             & epsilon decay strategy  & linear    & linear    & linear    \\
                             & batch size              & 32        & 32        & 32        \\
                             & hidden units            & 512       & 24        & 24        \\
\hline
\multirow{15}{*}{Elastic Step DQN}                & learning  rate        & 0.0001                       & 0.001                       & 0.001        \\
                             & target update horizon   & 250       & 20        & 250       \\
                             & replay capacity         & 10000     & 10000     & 10000     \\
                             & initial replay size     & 500       & 500       & 500       \\
                             & train frequency         & 1         & 1         & 1         \\
                             & gamma                   & 0.99      & 0.99      & 0.99      \\
                             & epsilon min             & 0.1       & 0.1       & 0.1       \\
                             & epsilon start           & 1         & 1         & 1         \\
                             & epsilon decay strategy  & linear    & linear    & linear    \\
                             & batch size              & 32        & 32        & 32        \\
                             & hidden units            & 200       & 24        & 24        \\
                             & alpha                   & 1         & 1         & 1         \\
                             & leaf size               & 40        & 40        & 40        \\
                             & minimum cluster size    & 5         & 5         & 5         \\
                             & metric                  & euclidean & euclidean & euclidean \\
                             & state memory bank                  & 10000 & 10000 & 1000 \\
                             & state memory batch size                  & 1000 & 1000 & 1000 \\
\hline                          
\caption{Final hyper-parameters used across all the experiments}
\end{longtable}

\end{document}